\documentclass[10pt,twocolumn,letterpaper]{article}

\usepackage{cvpr}              % To produce the CAMERA-READY version

\usepackage{xcolor}
\definecolor{iccvblue}{rgb}{0.21,0.49,0.74}
\definecolor{darkred}{RGB}{136, 8, 8}
\definecolor{darkgreen}{RGB}{9, 121, 105}
\definecolor{darkblue}{RGB}{0,0,180}
\definecolor{nbarrier}{RGB}{255, 120, 50}
\definecolor{nbicycle}{RGB}{255, 192, 203}
\definecolor{nbus}{RGB}{255, 255, 0}
\definecolor{ncar}{RGB}{0, 150, 245}
\definecolor{nconstruct}{RGB}{0, 255, 255}
\definecolor{nmotor}{RGB}{200, 180, 0}
\definecolor{npedestrian}{RGB}{255, 0, 0}
\definecolor{ntraffic}{RGB}{255, 240, 150}
\definecolor{ntrailer}{RGB}{135, 60, 0}
\definecolor{ntruck}{RGB}{160, 32, 240}
\definecolor{ndriveable}{RGB}{255, 0, 255}
\definecolor{nother}{RGB}{139, 137, 137}
\definecolor{nsidewalk}{RGB}{75, 0, 75}
\definecolor{nterrain}{RGB}{150, 240, 80}
\definecolor{nmanmade}{RGB}{213, 213, 213}
\definecolor{nvegetation}{RGB}{0, 175, 0}
\definecolor{nothers}{RGB}{0, 0, 0}
\usepackage{graphicx}
\usepackage{multirow}
\usepackage{makecell}
\usepackage{bm}
\usepackage{booktabs}
\usepackage{colortbl}
\usepackage{array}
\usepackage{amssymb}
\usepackage{pifont}
\usepackage{amsmath}
\usepackage{adjustbox}
\usepackage{subcaption} 
\usepackage{caption} 
\usepackage{float}
\usepackage{natbib}
\usepackage{wrapfig} % 导言区添加
\usepackage[utf8]{inputenc} % allow utf-8 input
\usepackage[T1]{fontenc}    % use 8-bit T1 fonts
\usepackage{url}            % simple URL typesetting
\usepackage{booktabs}       % professional-quality tables
\usepackage{amsfonts}       % blackboard math symbols
\usepackage{nicefrac}       % compact symbols for 1/2, etc.
\usepackage{microtype}      % microtypography
\usepackage{xcolor}         % colors
\usepackage{diagbox}

\definecolor{cvprblue}{rgb}{0.21,0.49,0.74}
\usepackage[pagebackref,breaklinks,colorlinks,allcolors=cvprblue]{hyperref}

\def\paperID{****} % *** 3958 Enter the Paper ID here
\def\confName{CVPR}
\def\confYear{2026}

\title{Test-Time 3D Occupancy Prediction}

\author{
Fengyi Zhang\textsuperscript{1}\;
Xiangyu Sun\textsuperscript{1}\;
Huitong Yang\textsuperscript{1}\;
Zheng Zhang\textsuperscript{2}\;
Zi Huang\textsuperscript{1}\;
Yadan Luo\textsuperscript{1}\thanks{Corresponding Author.}\\
{\textsuperscript{1}UQMM Lab, The University of Queensland} \;
{\textsuperscript{2}Harbin Institute of Technology}\\
}

\begin{document}
\maketitle
\begin{abstract}
  Self-supervised 3D occupancy prediction offers a promising solution for understanding complex driving scenes without requiring costly 3D annotations. However, training dense occupancy decoders to capture fine-grained geometry and semantics can demand \textit{hundreds} of GPU hours, and once trained, such models struggle to adapt to varying voxel resolutions or novel object categories without extensive retraining. To overcome these limitations, we propose a practical and flexible test-time occupancy prediction framework termed TT-Occ. Our method incrementally constructs, optimizes, and voxelizes time-aware 3D Gaussians from raw sensor streams by integrating vision foundation models (VFMs) at runtime. The flexible representation of 3D Gaussians enables voxelization at arbitrary user-specified resolutions, while the strong generalization capability of VFMs supports accurate perception and open-vocabulary recognition without requiring any network training or fine-tuning. To validate the generality and effectiveness of our framework, we present two variants: a LiDAR-based version and a vision-centric version, and conduct extensive experiments on the Occ3D-nuScenes and nuCraft benchmarks under varying voxel resolutions. Experimental results show that TT-Occ significantly outperforms existing computationally expensive pretrained self-supervised counterparts. Code is available at \url{https://github.com/Xian-Bei/TT-Occ}.  
% \href{https://github.com/Xian-Bei/TT-Occ}{Project page}.

  % Specifically, TT-Occ operates in a \textit{``lift-track-voxelize”} symphony: We first ``lift'' the geometry and semantics of surrounding-view extracted from VFMs to instantiate Gaussians at 3D space; Next, we ``track'' dynamic Gaussians while accumulating static ones to complete the scene and enforce temporal consistency; 
  % Finally, we voxelize the optimized Gaussians to generate occupancy prediction. 
  % Optionally, inherent noise in VFM predictions and tracking is mitigated by periodically smoothing neighboring Gaussians during optimization. 
  % Source code is available in the supplementary materials.

\end{abstract}

\begin{figure}[t]
    \centering
    \includegraphics[width=\linewidth]{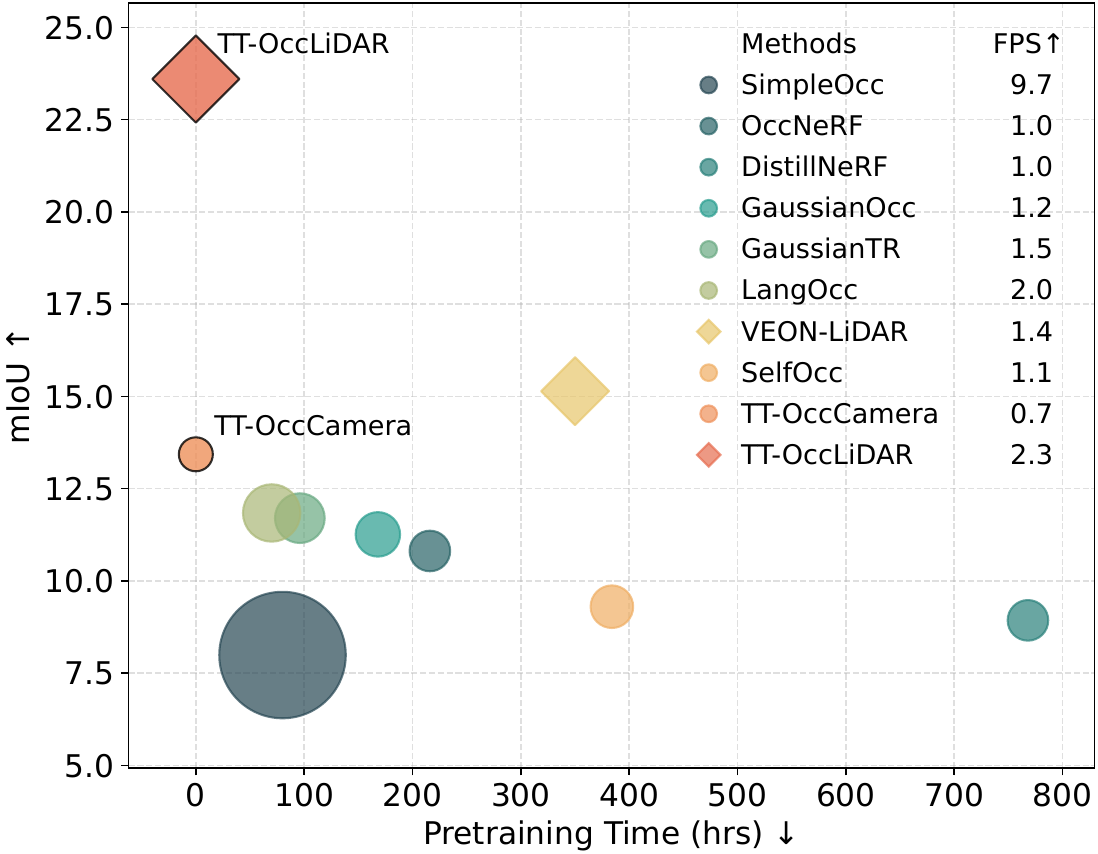}
    \caption{Comparison of self-supervised occupancy prediction methods in terms of pretraining time (x-axis), mIoU (y-axis), and runtime FPS (marker radius). TT-Occ achieves strong mIoU and competitive FPS without any pretraining.}
    \vspace{-2ex}
    \label{fig:scatter}
\end{figure}

\section{Introduction}
\label{sec:intro}
3D Occupancy prediction seeks to accurately identify regions within an environment that are occupied by objects of particular classes and those that remain free. This capability is crucial to enable collision-free trajectory planning and reliable navigation in autonomous driving systems \cite{planning_ad1,planning_ad2,3ddet,3ddet2} and embodied agents \cite{planning_agent,planning_agent2,navi,agent}.
Existing solutions \cite{TPVFormer,OccFormer,RenderOcc,OccFlowNet, SurroundOcc,GaussianFormer,Occ3D}, however, heavily rely on dense 3D annotations obtained through labor-intensive manual labeling of dynamic driving scenes spanning up to 80 meters per frame. To mitigate this cost, recent studies have resorted to \textit{self-supervised} alternatives \cite{SimpleOcc,OccNeRF,SelfOcc,DistillNeRF,GaussianOcc,GaussianTR,VEON,LangOcc,zhou2025autoocc}. While reducing labeling costs, they still require substantial computational overhead. For instance, training SelfOcc \cite{SelfOcc} on Occ3D-nuScenes \cite{Occ3D} at a voxel resolution of 0.4m requires approximately 2 days on eight GPUs. Furthermore, once trained, adapting to finer resolution (\textit{e.g.,} 0.2m of nuCraft \cite{nuCraft} dataset) or novel object classes (\textit{e.g.,} beyond the 17 predefined classes of Occ3D-nuScenes \cite{Occ3D}) may necessitate extensive retraining.

Recent vision foundation models (VFMs), however, change this landscape. Models such as VGGT \cite{vggt} and MapAnything \cite{keetha2025mapanything} provide reliable multiview geometry, and REX-Omni \cite{rexomni} enables open-vocabulary semantic reasoning, both directly at test time. As these cues become increasingly accessible without task-specific training, it is natural to revisit a long-standing assumption of the field: \textit{if such information no longer needs to be learned by a network, is training an occupancy model still necessary? }To answer this question, we explore a \underline{t}est-\underline{t}ime \underline{occ}upancy estimation method termed \textbf{TT-Occ}, which progressively constructs, optimizes and voxelizes time-aware 3D Gaussians from raw sensor streams by integrating VFMs. 
We introduce two variants, \textbf{TT-OccCamera} and \textbf{TT-OccLiDAR}, which differ in the sensor modality used to initialize the Gaussian primitives, respectively. Both variants eliminate costly pretraining and allow flexible adaptation to arbitrary voxel resolutions and user-specified semantic queries. Unlike previous NeRF- \cite{emernerf,unisim} and 3DGS-based \cite{DrivingGaussian,DriveStudio} reconstruction methods that perform offline per-scene modeling assisted by \textit{external GT priors} (\textit{e.g.,} HD maps or bounding boxes), TT-Occ generates occupancy representations in an online fashion, relying solely on raw sensor streams and generally trained VFMs to instantiate Gaussians capturing object geometry and semantics in unbounded outdoor scenes. 

Our framework follows a simple yet effective \textit{``lift-track-voxelize''} symphony: 
% \resizebox{1em}{1em}{\includegraphics{Figs/number_1.pdf}} 
(1) Lift: at each test time step, we first \textit{``lift''} geometry and semantic information of surrounding views extracted via VFMs into time-aware 3D Gaussians on the fly. The generated Gaussians can also be splatted back onto the image plane through differentiable rasterization for parameter optimization \cite{3DGS}. 
% \resizebox{1em}{1em}{\includegraphics{Figs/number_2.pdf}} 
(2) Track: next, we \textit{``track''} dynamic Gaussians and accumulate static ones using estimated motion flow. This motion compensates for partial object visibility and prevents trailing artifacts while maintaining long-term temporal coherence. 
% \resizebox{1em}{1em}{\includegraphics{Figs/number_3.pdf}} 
(3) Voxelize: at any given timestamp, the generated 3D Gaussians can be voxelized onto discrete occupancy grids with arbitrary user-specified resolutions. 
Optionally, to further mitigate the inherent noise in VFM predictions and tracking results, we introduce a Trilateral Radial Basis Function (TRBF), which jointly considers semantic, color, and spatial affinities to periodically smooth the Gaussian parameters.

Extensive experiments on Occ3D-nuScenes \cite{Occ3D} and the recently released high-resolution nuCraft \cite{nuCraft} demonstrate that TT-Occ achieves better performance than existing self-supervised counterparts, which typically require hundreds of GPU training hours. 
Qualitative analysis further highlights the superiority of TT-Occ in terms of temporal consistency and open-vocabulary generalization. 

\begin{figure*}
    \includegraphics[width=1\linewidth]{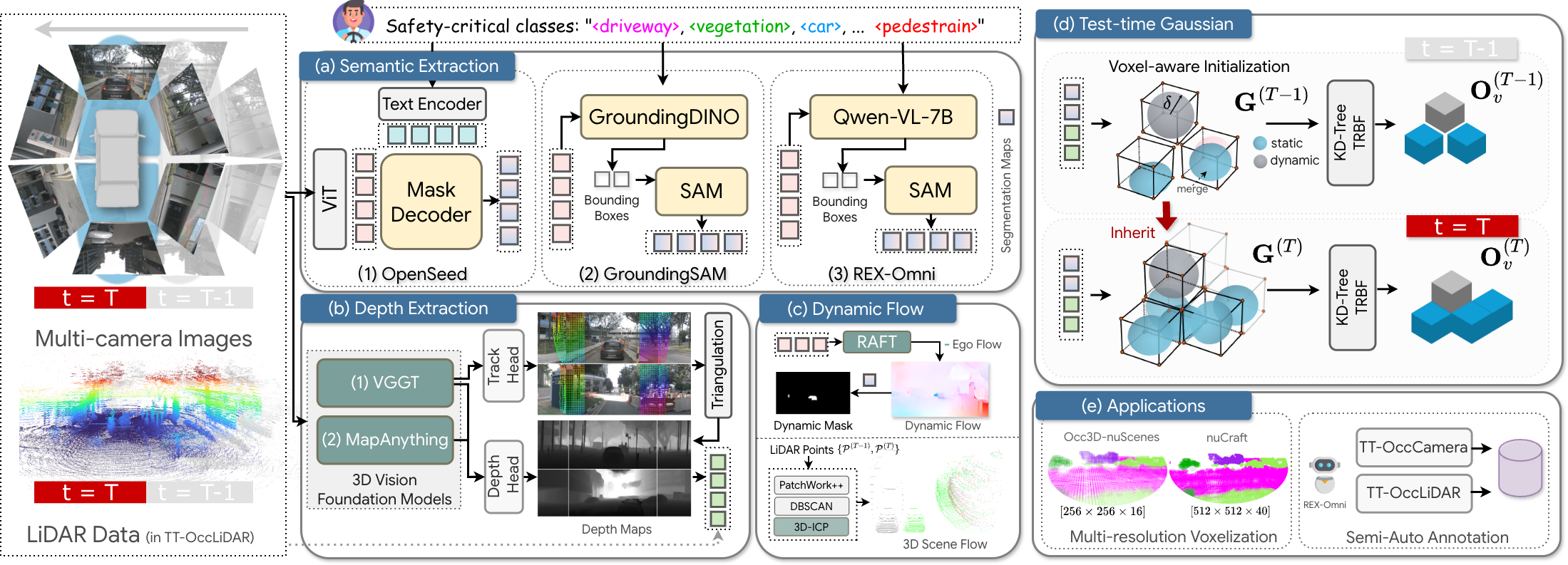}
    \caption{\textbf{Overview of the proposed TT-OccCamera and TT-OccLiDAR approaches.} TT-Occ performs test-time 3D occupancy estimation by directly integrating a suite of VFMs at runtime, avoiding any network training or fine-tuning.
(a) Multi-view semantics are obtained using arbitrary open-vocabulary segmentation VFMs (\textit{e.g.}, OpenSeed, GroundingSAM, REX-Omni).
(b) Geometry cues are extracted via depth/correspondence VFMs (\textit{e.g.}, VGGT, MapAnything).
(c) Dynamic flow is estimated to track moving objects and prevent trailing artifacts while accumulating static structure over time.
(d) The resulting features instantiate and refine time-aware 3D Gaussians, which can be voxelized at any resolutions for occupancy prediction.
(e) TT-Occ supports both LiDAR-based and camera-only variants and enables multi-resolution voxelization and semi-automatic annotation.
Despite leveraging multiple VFMs, TT-Occ remains highly efficient, delivering competitive occupancy performance across Occ3D-nuScenes and nuCraft.}\label{fig:overview}
\end{figure*}

\section{Related Work}
% \vspace{-1ex}
\label{sec:related work}

\noindent\textbf{Occupancy Prediction.} Fully supervised occupancy methods predict voxel-level semantics using dense voxel grids~\cite{TPVFormer,PanoOcc,HTCL}, depth priors~\cite{VoxFormer,Symphonize}, or sparse representations~\cite{Octreeocc,SparseOcc,GaussianFormer}. Despite their effectiveness, these approaches rely heavily on costly large-scale 3D annotations. To mitigate this, recent methods explore self-supervised occupancy learning. SelfOcc~\cite{SelfOcc} leverages signed distance fields (SDF) and multi-view stereo embeddings to achieve temporally consistent occupancy from videos. OccNeRF~\cite{OccNeRF} utilizes photometric consistency and 2D foundation model supervision for semantic occupancy estimation in unbounded scenes. In open-world scenarios, POP3D~\cite{POP-3D} jointly trains class-agnostic occupancy grids and open-vocabulary semantics using unlabeled paired LiDAR and images, but suffers from sparsity and semantic ambiguity due to low-resolution CLIP~\cite{CLIP} features. 
% VEON~\cite{VEON} uses LiDAR data for training, addressing depth ambiguities via enhanced depth models (MiDaS~\cite{MiDaS}, ZoeDepth~\cite{ZoeDepth}) and proposing high-resolution CLIP \cite{CLIP} embeddings. 
VEON \cite{VEON} introduces a vocabulary-enhanced occupancy framework trained with LiDAR supervision, leveraging CLIP features for open-vocabulary prediction and addressing depth ambiguities via enhance depth model (MiDaS~\cite{MiDaS}, ZoeDepth~\cite{ZoeDepth}).
% Gaussian-based methods, including GaussianFormer~\cite{GaussianFormer}, GaussianOcc~\cite{GaussianOcc}, and GaussianTR~\cite{GaussianTR}, utilize 3D Gaussian representations to efficiently model scenes. 
 % Specifically, GaussianFormer employs sparse Gaussian queries and cross-attention mechanisms for semantic voxel reconstruction,  while 
 GaussianOcc \cite{GaussianOcc} uses Gaussian Splatting \cite{3DGS} for cross-view optimization without pose annotations, while GaussianTR~\cite{GaussianTR} aligns rendered Gaussian features with pre-trained foundation models, enabling open-vocabulary occupancy prediction without explicit annotations.
Despite these advances, existing methods either rely on extensive offline training or struggle with open-vocabulary settings and fixed resolutions. In contrast, our approach overcomes these limitations by enabling occupancy prediction through temporally coherent, training-free Gaussian optimization at test time.

%-------------------------------------------------------------------------

% \noindent\textbf{Scene Modeling and Occupancy World Models.}
\noindent\textbf{3D Reconstruction of Driving Scenes.} Recent advances in dynamic scene modeling \cite{vggt,szymanowicz2025flash3d,keetha2025mapanything} have achieved impressive photorealism and multi-view consistency. OmniRe~\cite{DriveStudio} performs real-time 3D reconstruction and simulation by building local canonical spaces for dynamic urban actors. Street Gaussians~\cite{Street} separates moving vehicles from static backgrounds, enabling efficient and high-quality rendering. DrivingGaussian~\cite{DrivingGaussian} incrementally reconstructs static scenes and dynamically integrates moving objects via Gaussian graphs for interactive editing. HUGS~\cite{HUGS} jointly optimizes geometry, appearance, semantics, and motion to achieve real-time view synthesis and 3D semantic reconstruction without explicit bounding box annotations. Autoregressive world modeling methods~\cite{OccWorld,OccSora} predict future occupancy using previously estimated 3D occupancies, facilitating temporal reasoning in dynamic environments. Our test-time approach fundamentally \textit{differs} from these methods by eliminating dependencies on external priors and annotations (\textit{e.g.,} HD maps and GT bounding boxes). Instead, we focus solely on raw sensor inputs, optimizing Gaussian representations independently at each frame to directly infer the accurate geometry, rather than reconstructing photorealistic scenes or predicting future occupancy.

\section{Proposed Approach}

\noindent\textbf{Task Formulation.} At each time step $t$, the objective of occupancy estimation is to infer the voxelized geometry and semantic labels of the current scene directly from raw sensor inputs. Formally, we define the voxel grid as $\mathbf{O}^{(t)}\in\mathbb{C}^{\frac{X}{\delta}\times\frac{Y}{\delta}\times\frac{Z}{\delta}}$, where $X, Y, Z$ defines the spatial dimensions of the region of interest, and $\delta$ is the voxel resolution (\textit{e.g.,} 0.2m). 
The input modality varies by variant. The input of the vision-centric variant is $M$ surrounding-view camera images $\mathcal{I}^{(t)} = \{ \mathbf{I}^{(t)}_m \in \mathbb{R}^{3 \times H \times W}\}_{m=1}^{M}$, while LiDAR-based variant additionally takes a LiDAR point cloud $\mathcal{P}^{(t)} = \{ \mathbf{p}^{(t)}_i \in \mathbb{R}^3 \}_{i=1}^{N_t}$. Each voxel in $\mathbf{O}^{(t)}$ is assigned a semantic label from the set $\mathbb{C}=\{0, 1, \ldots, C\}$, where $0$ indicates an empty cell and labels $1$ to $C$ corresponds to distinct occupied categories.

\subsection{Lift Geometry and Semantics into Gaussians}
\label{sec:gaussian}
For each time step, a set of time-aware Gaussian blobs $\mathcal{G}^{(t)}=\{\mathbf{G}_i^{(t)}\}_{i=1}^{K_t}$ are instantiated to represent the scene. Each Gaussian is parameterized by its mean position $\boldsymbol{\mu}_i\in\mathbb{R}^3$, opacity $\alpha_i\in(0,1)$, color $\mathbf{c}_i\in\mathbb{R}^3$, semantic probability $\mathbf{m}_i\in\mathbb{R}^C$, time step $t$, and its spatial density is given by: 
\begin{equation}
    \mathbf{G}_i^{(t)}(\mathbf{x}) = \exp \left(-\frac{1}{2}(\mathbf{x} - \boldsymbol{\mu}_i)^\top \mathbf{\Sigma}_i^{-1}(\mathbf{x} - \boldsymbol{\mu}_i)\right),
\end{equation}
where covariance matrix $\mathbf{\Sigma}_i = R(\mathbf{q}_i)\operatorname{diag}(\mathbf{s}_i^2)R(\mathbf{q}_i)^\top$ is factorzied by the orientation quaternion $\mathbf{q}_i\in\mathbb{R}^4$ and the scale vector $\mathbf{s}_i\in\mathbb{R}_+^3$. 
To project each Gaussian on the 2D plane, we apply perspective transformation $\operatorname{Proj}(\mathbf{x}; \mathbf{K}, \mathbf{E})$ with the intrinsic matrix $\mathbf{K}\in\mathbb{R}^{3\times3}$ and extrinsic matrix $\mathbf{E}\in\mathbb{R}^{3\times4}$. The projected mean and covariance are:
\begin{equation}\nonumber
    \boldsymbol{\mu}^{\operatorname{2D}}_i = \operatorname{Proj}(\boldsymbol{\mu}_i)_{1:2},~ \mathbf{\Sigma}_i^{\operatorname{2D}} = \mathbf{J}_{\operatorname{Proj}}(\boldsymbol{\mu}_i)\Sigma_i \mathbf{J}_{\operatorname{Proj}}(\boldsymbol{\mu}_i)^{\top}_{1:2, 1:2}, 
\end{equation}
where $\mathbf{J}_{\operatorname{Proj}}$ is the Jacobian matrix. The color of the pixel $\mathbf{u}$ is then obtained by alpha blending.

\noindent \textbf{Modality-Specific Initialization.} In the LiDAR-based variant $\operatorname{TT-OccLiDAR}$, the sparse LiDAR points are directly initialized as 3D Gaussians, inheriting the precise spatial positions from real-world measurements. 
In contrast, the vision-centric variant $\operatorname{TT-OccCamera}$ reconstructs a 3D point cloud from 3D vision foundation models (3DVFMs). Specifically, we employ a pretrained 3DVFM such as VGGT \cite{vggt} and MapAnything \cite{keetha2025mapanything} to estimate dense depth maps from multi-view RGB inputs. When predicted depth maps suffer from inherent \textit{scale ambiguity} due to the lack of metric supervision (\textit{e.g.}, in VGGT), we mitigate this issue by applying multi-view triangulation to keypoint correspondences in overlapping views. See supplementary materials for details. 

\noindent \textbf{VFM Semantics.} To incorporate semantic information, we extract semantic maps from $M$ surrounding views by querying an open-vocabulary model \cite{OpenSeeD,groundedsam,rexomni}. Our framework is agnostic to the specific VFM used and various segmentation and grounding models were evaluated, as illustrated in Fig.~\ref{fig:overview}.
These semantic maps are then lifted to 3D via a visibility-weighted projection:
\begin{equation}\label{eq:semantic}
    \mathbf{m}_i = \frac{1}{M}\sum_{m=1}^M\mathbb{I}_{m}(\boldsymbol{\mu}_i)\mathcal{M}_m(\operatorname{Proj}\left(\boldsymbol{\mu}_i; \mathbf{K}_m, \mathbf{E}_m\right)),
\end{equation}
where $\mathbf{m}_i$ is a fused semantic probability vector for the 3D point. Specifically, $\boldsymbol{\mu_i} \in \mathbb{R}^3$ denotes the 3D coordinate of the $i$-th point. For each camera view $m \in \{1, \ldots, M\}$: $\mathrm{Proj}(\boldsymbol{\mu_i}; \mathbf{K}_m, \mathbf{E}_m)$ projects the 3D point $\boldsymbol{\mu_i}$ onto the 2D image plane using the corresponding camera intrinsics $\mathbf{K}_m$ and extrinsics $\mathbf{E}_m$; $\mathcal{M}_m(\cdot)$ returns the semantic probability vector from the open-vocabulary segmentation map at the projected 2D location; $\mathbb{I}_m(\boldsymbol{\mu_i})$ is a binary visibility indicator, equal to 1 if the point $\boldsymbol{\mu_i}$ is visible in view $m$, and 0 otherwise. The use of foundation models enables compatibility with open-vocabulary semantic queries, allowing TT-Occ to flexibly adapt to user-specified class definitions at test time.
For benchmark evaluation on nuScenes \cite{nuScenes}, we adopt the standard label space; however, our method inherently supports open-vocabulary settings without requiring retraining, in contrast to conventional self-supervised occupancy prediction approaches (\textit{e.g.,} \cite{SelfOcc}) that depend on fixed decoder architectures and label sets.

\noindent \textbf{Voxel-aware Simplifications.} To accelerate Gaussian optimization and subsequent voxelization, we simplify the standard 3DGS \cite{3DGS} by initializing the scale parameters with $\delta$ and constraining them using a sigmoid activation rather than an exponential function to prevent excessive growth.  Additionally, we prune redundant Gaussians within the same voxel cell (size $\delta$) while merging their semantic probabilities.

% ensuring the total number of Gaussians satisfies $K_t\ll N_t$. 

\begin{figure}
  \centering
  \includegraphics[width=1\linewidth]{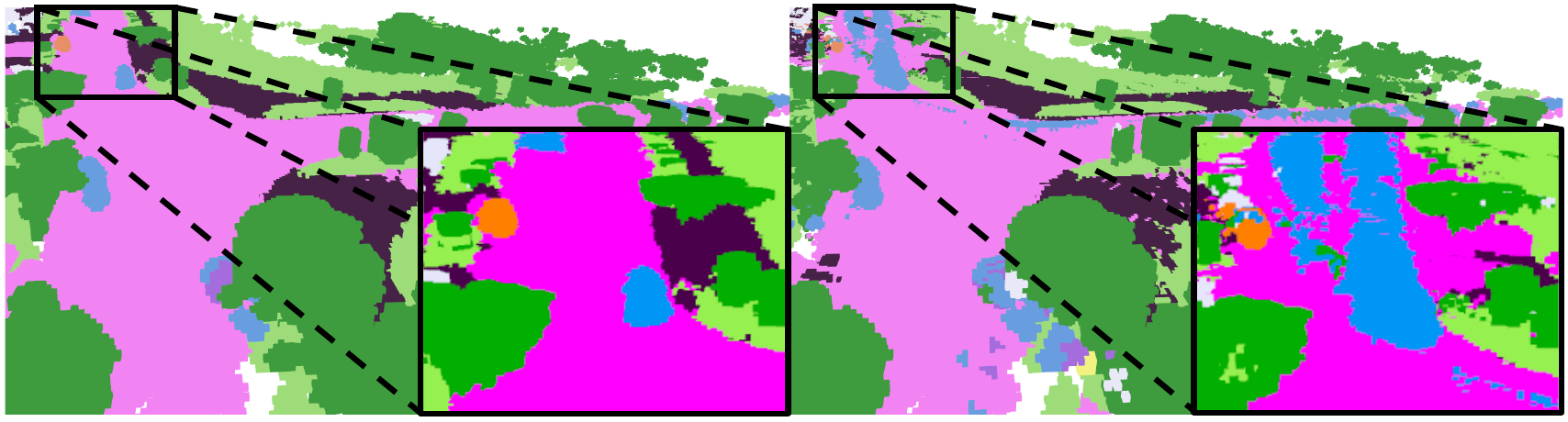}
  \caption{\small  \textbf{Illustration of trailing artifacts caused by naïvely accumulating per-frame Gaussians.}
Without handling dynamic regions, moving objects (\textit{e.g.}, cars shown in blue voxels) leave behind smeared or duplicated structures, corrupting the occupancy field. Our tracking suppresses these artifacts by separating dynamic Gaussians from static ones.}
  % \vspace{-2.5ex}
  \label{fig:trail}
\end{figure}

\subsection{Track Dynamic Gaussians}
\label{sec:flow}
Reconstructing a driving scene faithfully can be challenging due to fast-moving objects (\textit{e.g.,} vehicles, pedestrians) that are often only partially observed. Without prior knowledge such as complete trajectories or bounding box annotations of moving instances used in \cite{DrivingGaussian, DistillNeRF}, optimizing 3D Gaussians online can often result in severe \textit{trailing artifacts}. 
To address this, we propose to track dynamic Gaussians while accumulating static ones across adjacent frames.
In the upper image of Fig. \ref{fig:trail}, fast-moving vehicles (blue voxels) produce noticeable trailing artifacts, while the lower image shows a clean reconstruction after the Gaussian tracking.

\noindent \textbf{Modality-Specific Tracking.} Both $\operatorname{TT-OccCamera}$ and $\operatorname{TT-OccLiDAR}$ share the same mechanism for \textit{static} Gaussian inheritance, which enhances scene completeness by accumulating temporally consistent observations across frames. The key difference lies in how \textit{dynamic} Gaussians are handled, particularly in their tracking strategy and purpose. 
For $\operatorname{TT-OccLiDAR}$, we track the Gaussian motion with learning-free \textit{Gaussian scene flow} estimation, which allows us to relocate dynamic Gaussians accordingly. 
% Let $\boldsymbol{\mu}^{(t-1)}\in\mathcal{G}^{(t-1)}$ and $\boldsymbol{\mu}^{(t)}\in\mathcal{G}^{(t)}$ be the Gaussian centers at time steps $t-1$ and $t$ in the world coordinate system, respectively.
Our pipeline consists of the following steps: associating instances by projecting LiDAR points onto segmentation masks; denoising with DBSCAN \cite{dbscan}; matching clusters across frames based on spatial proximity and shape similarity; and finally estimating 3D flow using the iterative closest point (ICP) algorithm \cite{ICP}.
% Specifically, given a cluster of object Gaussians at the previous time step $\mathcal{C}^{(t-1)}\subset\boldsymbol{\mu}^{(t-1)}$ and the corresponding cluster at the current step $\mathcal{C}^{(t)}\subset\boldsymbol{\mu}^{(t)}$, we estimate the optimal rigid transformation $\mathbf{T}_{\mathcal{C}}  = [\mathbf{R}_{\mathcal{C}} , \mathbf{t}_{\mathcal{C}} ]$ by minimizing, 
% \vspace{-2ex}
% \begin{equation}\label{eq:flow}
    % \min_{\mathbf{R}_{\mathcal{C}}, \mathbf{t}_{\mathcal{C}}} \sum_{\mathbf{x}\in \mathcal{C}^{(t-1)}} \| \mathbf{R}_{\mathcal{C}}\mathbf{x} + \mathbf{t}_C - \operatorname{NN}(\mathbf{R}_C\mathbf{x} + \mathbf{t}_C, \mathcal{C}^{(t)})\|,
% \end{equation} 
% where $\operatorname{NN}(\cdot, \mathcal{C}^{(t)})$ returns the nearest neighbor points in $\mathcal{C}^{(t)}$. 
For $\operatorname{TT-OccCamera}$, we estimate optical flow between adjacent frames of the same camera using RAFT \cite{raft}, and compute ego-motion flow based on inter-frame camera poses and per-pixel depth predicted by 3DVFMs (\textit{e.g.}, VGGT \cite{vggt} or MapAnything \cite{keetha2025mapanything}). Subtracting the ego-motion flow from the optical flow yields a residual dynamic flow, which reflects true object motion. Although this 2D dynamic flow could, in principle, guide the 3D motion of dynamic Gaussians, back-projecting it into 3D space tends to amplify noise from both RAFT and 3DVFMs, resulting in unstable Gaussian motion.
To mitigate this, we adopt a compromise strategy by thresholding the dynamic flow magnitude to obtain a dynamic mask that identifies likely moving regions. The corresponding 3D Gaussians projected onto these regions are treated as dynamic and excluded from static accumulation in the next frame. While this approach does not allow accumulation of dynamic objects as in the LiDAR-based variant, it effectively reduces artifacts caused by noisy motion cues and temporal inconsistencies. 
See supplementary materials for implementation details.

\begin{table*}[t]
    \centering
    \caption{Occupancy prediction performance on \textbf{Occ3D-nuScenes} \cite{Occ3D}. 
    Best results among self-supervised methods are highlighted \textbf{in bold}. 
    % \vspace{-1ex} 
    }
    \renewcommand{\arraystretch}{1.3} % 调整行高
    \setlength{\tabcolsep}{4pt} % 缩小列间距
    \large % 增大字体
    \resizebox{\linewidth}{!}{
    \begin{tabular}{l c c c c ccccccccccccccc}
        \toprule
        $\operatorname{Method}$ & $\operatorname{Input}$ & $\operatorname{mIoU}$ $\uparrow$ 
        & \rotatebox{90}{\textcolor{nbarrier}{$\blacksquare$} $\operatorname{barr}$} %
    & \rotatebox{90}{\textcolor{nbicycle}{$\blacksquare$} $\operatorname{bike}$} %
    & \rotatebox{90}{\textcolor{nbus}{$\blacksquare$} $\operatorname{bus}$} %
    & \rotatebox{90}{\textcolor{ncar}{$\blacksquare$} $\operatorname{car}$} %
    & \rotatebox{90}{\textcolor{nconstruct}{$\blacksquare$} $\operatorname{c-veh}$} %
    & \rotatebox{90}{\textcolor{nmotor}{$\blacksquare$} $\operatorname{moto}$} %
    & \rotatebox{90}{\textcolor{npedestrian}{$\blacksquare$} $\operatorname{ped}$} %
    & \rotatebox{90}{\textcolor{ntraffic}{$\blacksquare$} $\operatorname{t-cone}$} %
    & \rotatebox{90}{\textcolor{ntrailer}{$\blacksquare$} $\operatorname{trail}$} %
    & \rotatebox{90}{\textcolor{ntruck}{$\blacksquare$} $\operatorname{truck}$} %
    & \rotatebox{90}{\textcolor{ndriveable}{$\blacksquare$} $\operatorname{d-surf}$} %
    & \rotatebox{90}{\textcolor{nsidewalk}{$\blacksquare$} $\operatorname{s-walk}$} %
    & \rotatebox{90}{\textcolor{nterrain}{$\blacksquare$} $\operatorname{terr}$} %
    & \rotatebox{90}{\textcolor{nmanmade}{$\blacksquare$} $\operatorname{man}$} %
    & \rotatebox{90}{\textcolor{nvegetation}{$\blacksquare$} $\operatorname{vege}$} \\ 
        \midrule
        $\operatorname{BEVFormer}_{\textcolor{gray}{\small\operatorname{(ECCV'22)}}}$ \cite{BEVFormer}& C & 26.88 & 37.83 & 17.87 & 40.44 & 42.43 & 7.36 & 23.88 & 21.81 & 20.98 & 22.38 & 30.70 & 55.35 & 36.0 & 28.06 & 20.04 & 17.69 \\ 
        $\operatorname{CTF-Occ_{\textcolor{gray}{\small\operatorname{(NeurIPS'23)}}}}$ \cite{Occ3D} & C & 28.53 & 39.33 & 20.56 & 38.29 & 42.24 & 16.93 & 24.52 & 22.72 & 21.05 & 22.98 & 31.11 & 53.33 & 37.98 & 33.23 & 20.79 & 18.0\\ 
        \midrule
        $\operatorname{RenderOcc}_{\textcolor{gray}{\small\operatorname{(ICRA'24)}}}$ \cite{RenderOcc}& C & 23.93 & 27.56 & 14.36 & 19.91 & 20.56 & 11.96 & 12.42 & 12.14 & 14.34 & 20.81 & 18.94 & 68.85 & 42.01 & 43.94 & 17.36 & 22.61\\
        $\operatorname{OccFlowNet}_{\textcolor{gray}{\small\operatorname{(WACV'25)}}}$ \cite{OccFlowNet}& C\&L & 26.14 & 27.50 & 26.00 & 34.00 & 32.00 & 20.40 & 25.90 & 18.60 & 20.20 & 26.00 & 28.70 & 62.00 & 37.80 & 39.50 & 29.00 & 26.80 \\ 
        \midrule
        $\operatorname{SimpleOcc}_{\textcolor{gray}{\small\operatorname{(TIV'24)}}}$ \cite{SimpleOcc}& C & 7.99 & 0.67 & 1.18 & 3.21 & 7.63 & 1.02 & 0.26 & 1.80 & 0.26 & 1.07 & 2.81 & 40.44 & 18.30 & 17.01 & 13.42 & 10.84\\ 
        $\operatorname{OccNeRF}_{\textcolor{gray}{\small\operatorname{(Arxiv'24)}}}$ \cite{OccNeRF}& C & 10.81 & 0.83 & 0.82 & 5.13 & 12.49 & 3.50 & 0.23 & 3.10 & 1.84 & 0.52 & 3.90 & 52.62 & 20.81 & 24.75 & 18.45 & 13.19\\ 
        $\operatorname{DistillNeRF}_{\textcolor{gray}{\small\operatorname{(NeurIPS'24)}}}$ \cite{DistillNeRF}& C & 8.93 & 1.35 & 2.08 & 10.21 & 10.09 & 2.56 & 1.98 & 5.54 & 4.62 & 1.43 & 7.90 & 43.02 & 16.86 & 15.02 & 14.06 & 15.06 \\ 
        $\operatorname{GaussianOcc}_{\textcolor{gray}{\small\operatorname{(Arxiv'24)}}}$ \cite{GaussianOcc}& C & 11.26 & 1.79 & 5.82 & 14.58 & 13.55 & 1.30 & 2.82 & 7.95 & 9.76 & 0.56 & 9.61 & 44.59 & 20.10 & 17.58 & 8.61 & 10.29\\
        $\operatorname{GaussianTR}_{\textcolor{gray}{\small\operatorname{(CVPR'25)}}}$ \cite{GaussianTR}& C & 11.70 & 2.09 & 5.22 & 14.07 & 20.43 & 5.70 & 7.08 & 5.12 & 3.93 & 0.92 & 13.36 & 39.44 & 15.68 & 22.89 & 21.17 & 21.87\\ 
 
        $\operatorname{LangOcc}_{\textcolor{gray}{\small\operatorname{(3DV'25)}}}$ \cite{LangOcc}& C & 11.84 & 3.10 & 9.00 & 6.30 & 14.20 & 0.40 & 10.80 & 6.20 & 9.00 & 3.80 & 10.70 & 43.70 & 9.50 & 26.40 & 19.60 & 26.40 \\ 
        $\operatorname{VEON-LiDAR}_{\textcolor{gray}{\small\operatorname{(ECCV'24)}}}$ \cite{VEON}& C\&L & 15.14 & 10.40 & 6.20 & \textbf{17.70} & 12.70 & 8.50 & 7.60 & 6.50 & 5.50 & 8.20 & 11.80 & 54.50 & 25.50 & 30.20 & 25.40 & 25.40\\
        $\operatorname{SelfOcc}_{\textcolor{gray}{\small\operatorname{(CVPR'24)}}}$ \cite{SelfOcc} & C & 10.54 & 0.15 & 0.66 & 5.46 & 12.54 & 0.00 & 0.80 & 2.10 & 0.00 & 0.00 & 8.25 & \textbf{55.49} & 26.30 & 26.54 & 14.22 & 5.60\\ 
        \midrule
        % \rowcolor{blue!5} $\operatorname{TT-OccCamera}$ & C & 13.43 & 0.00 & 5.90 & 8.94 & 12.58 & 2.75 & 9.67 & 4.71 & 4.04 & 0.00 & 8.77 & \textbf{55.65} & 26.49 & 30.20 & 15.13 & 16.57\\
        % \rowcolor{blue!5} $\operatorname{TT-OccLiDAR}$ & C\&L & \textbf{23.60} & 0.00 & \textbf{15.99} & \textbf{23.01} & \textbf{25.42} & 5.61 & \textbf{20.50} & \textbf{20.68} & 7.36 & 0.00 & \textbf{24.32} & 51.89 & \textbf{31.06} & \textbf{37.15} & \textbf{43.87} & \textbf{47.20}\\

        \rowcolor{blue!5} $\operatorname{TT-OccCamera}$ & C & 16.70 & 21.51 & 10.46 & 10.70 & 14.72 & 11.89 & 12.33 & 9.71 & 12.20 & 4.37 & 7.92 & 48.27 & 23.70 & 28.31 & 14.13 & 20.24\\
        \rowcolor{blue!5} $\operatorname{TT-OccLiDAR}$ & C\&L & \textbf{27.41} & \textbf{32.44} & \textbf{16.08} & 16.88 & \textbf{22.56} & \textbf{28.32} & \textbf{21.06} & \textbf{14.34} & \textbf{20.02} & \textbf{10.33} & \textbf{14.47} & 52.60 & \textbf{31.40} & \textbf{37.21} & \textbf{48.02} & {45.37}\\
        \bottomrule
    \end{tabular}}
    \label{tab:occ3d}
\end{table*}

\subsection{Gaussian Voxelization}
\label{sec:fusion}
% The loss function for each test frame $t$ combines
% \begin{equation}\nonumber
%     \mathcal{L} = \sum_{\mathbf{u}\in\mathbf{I}^{(t)}_m}|\hat{\mathbf{I}}^{(t)}_m(\mathbf{u}) - \mathbf{I}_m^{(t)}(\mathbf{u})| +\lambda \sum_{\mathbf{v}\in\mathbf{p}_{\operatorname{2D}, m}^{(t)}}|\hat{\mathbf{D}}^{(t)}(\mathbf{v}) - \mathbf{D}^{(t)}(\mathbf{v})|,
% \end{equation}
% where the coefficient $\lambda$ balances the photometric consistency and depth consistency loss. $\mathbf{p}_{\operatorname{2D}, m}^{(t)}$ denotes the 2D projection of LiDAR points onto the $m$-th view.
% , direct optimization can amplify and propagate errors across neighboring Gaussians. 
% We define Trilateral RBF Kernel to smooth the evolving Gaussian set $\mathcal{G}^{(t)}$. 
% Unlike uniform aggregation, our TRBF kernel leverages the covariance structure of 3D Gaussians to guide anisotropic information propagation, preserving local object structures and semantic boundaries. 
Following 3DGS \cite{3DGS}, our model refines Gaussian parameters at test time by minimizing a loss that enforces color consistency, with sky regions intentionally masked out as in \cite{DrivingGaussian}. 
Optionally, to further mitigate the inherent noise and errors in VFMs' predictions and tracking results, we introduce a Trilateral Radial Basis Function (TRBF) kernel for periodic smoothing and denoising.
TRBF kernel improves the spatial and temporal coherence of occupancy predictions by leveraging spatial, radiometric, and semantic affinities among Gaussians for anisotropic information propagation while preserving local structures and semantic boundaries.
Formally, for each $\mathbf{m}_i\in\mathbf{G}_i^{(t)}$, the kernel smoothing is defined as a deformable convolution over its nearest neighbors:
\begin{equation}
\label{eq:filter}
\mathbf{m}_i \leftarrow \frac{1}{Z(i)} \sum_{j \in \operatorname{NN}(i)} \mathbf{m}_j \cdot \mathcal{K}(i, j),
\end{equation}
where $\operatorname{NN}(\cdot)$ identifies $K$ nearest  Gaussians using a  KD-Tree for efficient search and $Z(i)$
is a normalization factor $Z(i)=\sum_{j\in \operatorname{NN(i)}} \mathcal{K}(i, j)$ to ensure that $\mathbf{m}_i$ sums to 1 as a valid probability. By the Schur Product Theorem, the trilateral kernel decomposes element-wise into spatial, radiometric, and semantic components: 
\begin{equation}
\mathcal{K}(i, j) = \mathcal{K}_{\boldsymbol{\mu}}(i, j) \cdot \mathcal{K}_{\mathbf{c}}(i, j) \cdot \mathcal{K}_{\mathbf{m}}(i, j),
\end{equation}
where each term $\operatorname{attr} \in \{\boldsymbol{\mu}, \mathbf{c}, \mathbf{m}\} $ is defined as the following format:
\begin{equation}
\begin{split}
    &\mathcal{K}_{\operatorname{attr}}(i, j) = \exp (-\frac{\|\operatorname{attr}_i - \operatorname{attr}_j\|^2}{2\sigma_{\operatorname{attr}}^2} ).
\end{split}
\end{equation}
From a signal processing perspective, the trilateral smoothing behaves as a \textit{non-stationary low-pass} filter with locally adaptive cutoff frequencies. 
% Specially, each term is defined as,
% \begin{equation}\nonumber
% \begin{split}
%     &\mathcal{K}_{\operatorname{spatial}} = \exp ( -\frac{(\boldsymbol{\mu}_i - \boldsymbol{\mu}_j)^\top \mathbf{\Sigma}_{ij}^{-1} (\boldsymbol{\mu}_i - \boldsymbol{\mu}_j)}{2\sigma_{\mu}^2} ),\\
%     &\mathcal{K}_{\operatorname{radio}} = \exp (-\frac{\|\mathbf{c}_i - \mathbf{c}_j\|^2}{2\sigma_c^2} ),\mathcal{K}_{\operatorname{sem}}=\exp ( -\frac{D_{\text{KL}}(\mathbf{m}_i \| \mathbf{m}_j)}{2\sigma_s^2} ),
% \end{split}
% \end{equation}
% where $\mathbf{\Sigma}_{ij} = (\mathbf{\Sigma}_i^{-1} + \mathbf{\Sigma}_j^{-1})^{-1}$ is the fused covariance matrix, capturing directional uncertainty and structural alignment.  

For efficient occupancy estimation, we voxelize the accumulated Gaussians $\mathcal{G}^{(t)}$
% $\mathbf{V}^{(t)}=\mathcal{G}^{(t)}\cup\mathcal{H}_{\operatorname{static}}$. 
into a discrete grid $\mathbf{\Omega}=[\frac{X}{\delta}\times \frac{Y}{\delta} \times \frac{Z}{\delta}]$, 
% Instead of uniform contributions, 
where each Gaussian's contribution on a voxel is weighted based on its spatial proximity. 
Formally, the semantic probability of a voxel $v\in\mathbf{\Omega}$ is given by:
\begin{equation}
\label{eq:voxelization}
\begin{split}
       &\mathbb{P}(\mathbf{O}_v^{(t)}) =
\frac{1}{Z_v} \sum_{\mathbf{G}_i^{(t)}\in\mathcal{G}^{(t)}}
\left( \mathbf{m}_i \cdot \mathcal{K}_{\boldsymbol{\mu}}(i, v) \right),
\end{split}
\end{equation}
where $Z_v$ is the normalizing factor to ensure that $\mathbb{P}(\mathbf{O}_v^{(t)})$ sums to 1 as a valid probability. This voxelization strategy allows flexible scaling to varying voxel resolutions during test-time, balancing efficiency and precision.

\begin{table*}[ht!]
    \centering
    \caption{3D occupancy prediction performance on the high-resolution \textbf{nuCraft} dataset \cite{nuCraft}. Pretraining time is reported in GPU hours.
    % As no prior self-supervised methods have been trained or evaluated in this setting, we modify $\operatorname{SelfOcc}$ \cite{SelfOcc} using its official code as our baseline.
    % \vspace{-1ex}
    }
    \renewcommand{\arraystretch}{1.2}
    \setlength{\tabcolsep}{6pt}
    \Large
    \resizebox{\linewidth}{!}{
    \begin{tabular}{l  c  c  c c  c c c c c c c c c c c c c c}
        \toprule
       $\operatorname{Method}$ & $\operatorname{3D~GT}$  & $\operatorname{Pretraining}$ & $\operatorname{mIoU}$ $\uparrow$ 
        & \rotatebox{90}{\textcolor{nbarrier}{$\blacksquare$} $\operatorname{barr}$} %
    & \rotatebox{90}{\textcolor{nbicycle}{$\blacksquare$} $\operatorname{bike}$} %
    & \rotatebox{90}{\textcolor{nbus}{$\blacksquare$} $\operatorname{bus}$} %
    & \rotatebox{90}{\textcolor{ncar}{$\blacksquare$} $\operatorname{car}$} %
    & \rotatebox{90}{\textcolor{nconstruct}{$\blacksquare$} $\operatorname{c-veh}$} %
    & \rotatebox{90}{\textcolor{nmotor}{$\blacksquare$} $\operatorname{moto}$} %
    & \rotatebox{90}{\textcolor{npedestrian}{$\blacksquare$} $\operatorname{ped}$} %
    & \rotatebox{90}{\textcolor{ntraffic}{$\blacksquare$} $\operatorname{t-cone}$} %
    & \rotatebox{90}{\textcolor{ntrailer}{$\blacksquare$} $\operatorname{trail}$} %
    & \rotatebox{90}{\textcolor{ntruck}{$\blacksquare$} $\operatorname{truck}$} %
    & \rotatebox{90}{\textcolor{ndriveable}{$\blacksquare$} $\operatorname{d-surf}$} %
    & \rotatebox{90}{\textcolor{nsidewalk}{$\blacksquare$} $\operatorname{s-walk}$} %
    & \rotatebox{90}{\textcolor{nterrain}{$\blacksquare$} $\operatorname{terr}$} %
    & \rotatebox{90}{\textcolor{nmanmade}{$\blacksquare$} $\operatorname{man}$} %
    & \rotatebox{90}{\textcolor{nvegetation}{$\blacksquare$} $\operatorname{vege}$}  \\ %
        % \midrule
        % M-CONet & Dense 3D & 20.7 & 24.4 & 14.4 & 22.3 & 25.4 & 16.4 & 17.0 & 19.1 & 14.4 & 16.4 & 21.8 & 34.3 & - & - & - &22.1\\
        % nuCraft \cite{nuCraft} & Dense 3D & 26.2 & 27.6 & 21.6 & 23.7 & 28.8 & 15.5 & 29.2 & 34.3 & 20.5 & 19.2 & 24.0 & 40.1 & - & - & - & 28.6\\
        \midrule
        $\operatorname{C-CONet}_{\textcolor{gray}{\small\operatorname{(ICCV'23)}}}$ \cite{C-CONet}
        % & Raw Sensor
        & Dense & - & 13.4  & 14.30 & 9.10 & 16.50 & 18.30 & 7.40 & 12.30 & 11.10 & 9.40 & 5.80 & 13.20 & 32.50 & - & - & - & 19.90\\ 
        \midrule
        % 01:50:59 - 00:00:13 = 01:50:26 = 110*60+26 = 6626 / 6019
        $\operatorname{SelfOcc}^{\dagger}_{\textcolor{gray}{\small\operatorname{(CVPR'24)}}}$ \cite{SelfOcc} 
        % & Raw Sensor
        &\textcolor{darkgreen}{\ding{55}} & 384 hrs & 2.22 & 0.41 & 0.54 & 2.79 & 7.12 & 0.00 & 0.81 & 1.67 & 0.00 & 0.00 & 5.50 & 2.41 & 3.88 & 3.55 & 1.96 & 2.72\\ 
        % GaussianOcc$^\dagger$ \cite{GaussianOcc} 
        % & ~ & ~ & ~ & ~ & ~ & ~ & ~ & ~ & ~ & ~ & ~ & ~ & ~ & ~ & ~ & ~\\ 
        \rowcolor{blue!5} $\operatorname{TT-OccCamera}$
        % & Raw Sensor 
        &\textcolor{darkgreen}{\ding{55}}&\textcolor{darkgreen}{\ding{55}}& 5.95 & 9.21 & 5.58 & 5.35 & 6.19 & 3.93 & 7.28 & 4.86 & 4.73 & 2.89 & 3.44 & 10.15 & 6.17 & 7.18 & 3.92 & 8.45 \\
        % \rowcolor{gray!15} w/. Ours-M & $C\&L$ & \textbf{35.4}\textcolor{red}{\scriptsize +5.9} & \textbf{23.9}\textcolor{red}{\scriptsize +3.8} \\
        \rowcolor{blue!5} $\operatorname{TT-OccLiDAR}$
        % & Raw Sensor 
        &\textcolor{darkgreen}{\ding{55}}&\textcolor{darkgreen}{\ding{55}}& \textbf{10.92} & \textbf{14.92} & \textbf{8.15} & \textbf{12.21} & \textbf{11.93} & \textbf{10.71} & \textbf{13.82} & \textbf{6.25} & \textbf{8.39} & \textbf{6.93} & \textbf{9.05} & \textbf{12.57} & \textbf{8.41} & \textbf{10.29} & \textbf{12.89} & \textbf{17.37} \\
        % \rowcolor{gray!15} w/. Ours-M & $C\&L$ & \textbf{35.4}\textcolor{red}{\scriptsize +5.9} & \textbf{23.9}\textcolor{red}{\scriptsize +3.8} \\
        \bottomrule
    \end{tabular}}
    \label{tab:nu}
    % \vspace{-1ex}
\end{table*}
\section{Experiments}

\subsection{Experiment Setup}  

% \subsubsection{Datasets and Metrics}  
% \noindent\textbf{Datasets.} 
% To evaluate the versatility of TT-Occ, 
Experiments were conducted on the widely used nuScenes \cite{nuScenes} benchmark using 3D occupancy GT from {Occ3D-nuScenes} \cite{Occ3D} and {nuCraft} \cite{nuCraft}. 
The nuScenes dataset consists of 600 training scenes and 150 validation ones.
% , each representing a 20-second driving sequence captured at 2 Hz.
% by six surround-view cameras and a top-mounted 32-beam LiDAR. 
Existing supervised and self-supervised methods typically require extensive offline training on the training split. 
In contrast, TT-Occ requires no pretraining and is directly evaluated on the validation split.
% , demonstrating its ability to generalize without reliance on labeled training data. 
In particular, {Occ3D-nuScenes} \cite{Occ3D} provides voxelized occupancy annotations at \textit{0.4m} resolution, covering a spatial range o \([-40m, 40m]\) along the X and Y axes and \([-1m, 5.4m]\) along the Z axis. 
{nuCraft} \cite{nuCraft} offers more finer-grained annotations with a resolution of \textit{0.2m}, covering \([-51.2m, 51.2m]\) in the X and Y directions and \([-5m, 3m]\) in the Z direction.  
% Compared to offline-trained models, which require a predefined class set before training, our test-time approach generalizes dynamically to arbitrary object classes without retraining. 
% For a fair comparison, we follow standard evaluation protocols, reporting results on 17 predefined classes, while open-vocabulary generalization is explored separately in \cref{exp:main}.
% \noindent\textbf{Evaluation Metrics.} 

We evaluate semantic occupancy prediction using mean Intersection over Union ($\operatorname{mIoU}$), computed as the average $\operatorname{IoU}$ across all classes. Following prior works \cite{OccNeRF,GaussianOcc,GaussianTR}, we exclude the ``noise” and ``other flat” categories, as these do not correspond to valid prompts in open-vocabulary segmentation. 
% Additionally, we report inference speed and memory consumption (VRAM) to provide a comprehensive comparison with SelfOcc \cite{SelfOcc}.
% \subsection{Baselines}
We primarily compare our method with self-supervised counterparts, including SimpleOcc \cite{SimpleOcc}, OccNeRF \cite{OccNeRF}, DistillNeRF \cite{DistillNeRF}, GaussianOcc \cite{GaussianOcc}, GaussianTR \cite{GaussianTR}, LangOcc \cite{LangOcc}, VEON \cite{VEON}, and SelfOcc \cite{SelfOcc}. 
These methods represent a broad range of self-supervised occupancy research and include both NeRF \cite{NeRF} and 3DGS \cite{3DGS} representation.
For reference, we also include results from self-supervised methods that serve as upper bounds for performance comparison. These include BEVFormer \cite{BEVFormer} and CTF-Occ \cite{Occ3D}, which are trained with dense 3D voxel-level annotations, as well as RenderOcc \cite{RenderOcc} and OccFlowNet \cite{OccFlowNet}, which are trained on sparse point-level ground truth.
All experiments were conducted on an NVIDIA RTX 6000 Ada (48GB) GPU.
Please refer to the supplementary materials for implementation details.
% Implementation details are provided in the supplementary material.
% , offering a meaningful basis for comparison.  
% Additionally, we include two representative methods for each supervision type to highlight the development of self-supervised techniques. 
% Specifically, 

% We initialize the 3D Gaussian scale at \(0.2\), simply leveraging the minimal resolution of the evaluated dataset (nuCraft \cite{nuCraft}) as a reference.  
% Although additional iterations can further improve accuracy, our initialization already provides a strong baseline and is well-suited for scenarios with strict speed requirements.  
% This choice accounts for the inherent noise in semantic information obtained from VLM predictions, whereas raw sensor images are generally more reliable.  

\begin{figure*}[ht!]
  \centering
  % \fbox{\rule{0pt}{6in} \rule{\linewidth}{0pt}}
   \includegraphics[width=\linewidth]{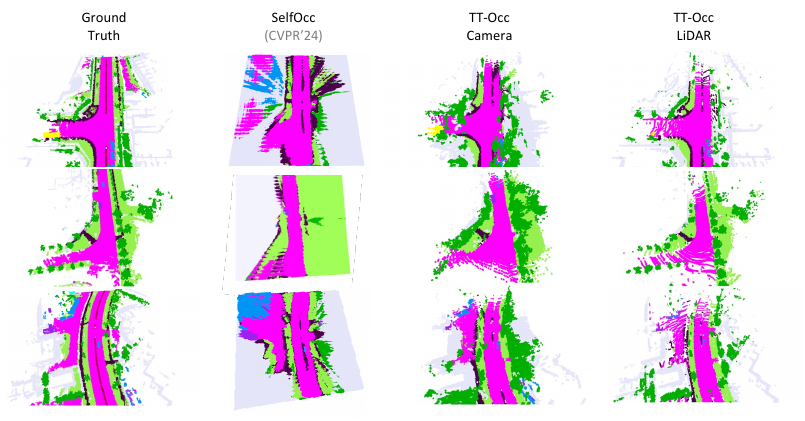}   
   % \vspace{-5ex}
   \caption{Qualitative comparisons on nuCraft \cite{nuCraft} between both variants of the proposed TT-Occ and SelfOcc \cite{SelfOcc}.
   % Our approach more accurately reconstructs scene geometry, preserving finer details and capturing object structures with higher fidelity. 
   % More visualizations can be found in the supplementary material.
   }
   \label{fig:qualitative}
   % \vspace{-2ex}
\end{figure*}

\begin{table}[!htb]
\centering
\setlength{\tabcolsep}{3.8pt} 
\renewcommand{\arraystretch}{1.05}
\caption{Comparison of TT-Occ and SelfOcc \cite{SelfOcc} using the same OpenSeed \cite{OpenSeeD} VFM under different RayIoU thresholds. }
\resizebox{\columnwidth}{!}{
\begin{tabular}{lcccc}
\toprule
\textbf{Method} & \textbf{mIoU} & \textbf{RayIoU@1} & \textbf{RayIoU@2} & \textbf{RayIoU@4} \\
\midrule
$\operatorname{SelfOcc}$ \cite{SelfOcc}& 10.5 & 8.9 & 10.5 & 12.0 \\
$\operatorname{TT-OccCamera}$ & 13.4& 10.0 & 12.9 & 15.7 \\
$\operatorname{TT-OccLiDAR}$ & \textbf{23.6} & \textbf{20.9} & \textbf{24.1} & \textbf{25.8} \\
\bottomrule
\end{tabular}}
\label{tab:rayiou}
% \vspace{-2.5ex}
\end{table}
\subsection{Main Results}\label{exp:main}
% \subsubsection{Quantitative Comparisons}
% \vspace{-1ex}

\noindent\textbf{Results on Occ3D-nuScenes} are shown in Table \ref{tab:occ3d}.
It is evident that \textit{both} variants of TT-Occ not only eliminate the need for costly offline training but also surpass the previous SOTA. 
Notably, $\operatorname{TT\text{-}OccLiDAR}$ even achieves an mIoU of 27.41, \textit{surpassing} RenderOcc \cite{RenderOcc} (23.93), which is trained with \textit{sparse 3D ground truth}.
Moreover, our camera-only variant $\operatorname{TT\text{-}OccCamera}$ reaches an mIoU of 16.70, \textit{exceeding} VEON-LiDAR \cite{VEON} (15.14), which is trained using \textit{LiDAR supervision}.
In addition, compared to the SOTA self-supervised baseline SelfOcc \cite{SelfOcc}, our approach achieves higher IoU not only for frequently occurring, large-area categories such as terrain and vegetation, but also shows substantial improvements on rare, dynamic, and small-area categories such as motorcycle, bus, and pedestrian.

To further evaluate the geometric quality beyond conventional mIoU, we report RayIoU under multiple thresholds, following the standard protocol in \cite{rayiou}. Importantly, all methods are equipped with the same OpenSeed \cite{OpenSeeD} VFM for fair comparison. As shown in Table \ref{tab:rayiou}, both TT-Occ variants achieve consistent gains across all RayIoU levels, indicating that our time-aware 3D Gaussians improve not only occupancy classification but also the metric accuracy of reconstructed geometry along camera rays. TT-OccCamera improves the RayIoU@4 metric by 30.8\% over SelfOcc, while TT-OccLiDAR delivers the strongest performance with a substantial 115\% improvement. These results highlight that TT-Occ suppresses temporal drift and produces cleaner free-space estimates, where those qualities that are not fully captured by mIoU alone. In practice, higher RayIoU translates to more reliable scene geometry for downstream planning and safety-critical perception modules, underscoring the superiority of TT-Occ given the same semantic VFM.

\begin{figure*}[t]
\centering
\setlength{\tabcolsep}{3pt}

\begin{minipage}[t]{0.49\textwidth}
    \centering
    
    \includegraphics[width=\linewidth]{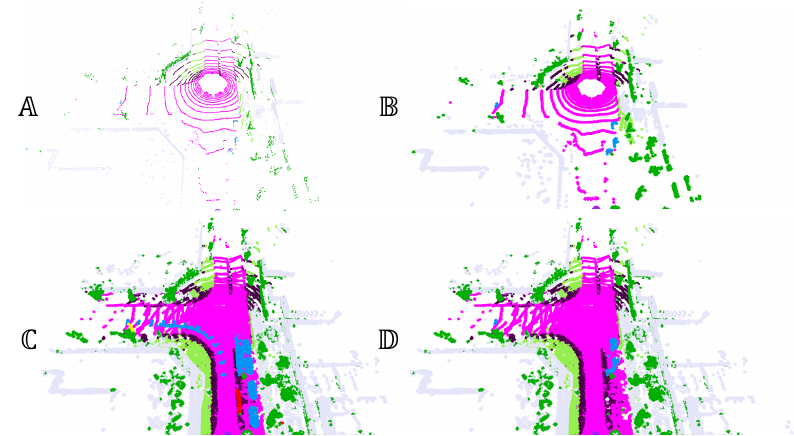}
    \caption*{(a) Ablation Results of $\operatorname{TT\text{-}OccLiDAR}$ }
    % \captionof{table}{Ablatio}
    \setlength{\tabcolsep}{7pt} 
    \resizebox{0.85\linewidth}{!}{\hbox to 1.0\linewidth{
    \begin{tabular}{c|c c c c}
        \toprule
        No. & mIoU~$\uparrow$
        & \rotatebox{90}{\textcolor{nbus}{$\blacksquare$} $\operatorname{bus}$}
        & \rotatebox{90}{\textcolor{npedestrian}{$\blacksquare$} $\operatorname{ped}$}
        & \rotatebox{90}{\textcolor{nmanmade}{$\blacksquare$} $\operatorname{man}$}\\
        \midrule
        $\mathbb{A}$ & 7.3 & 5.0 & 9.8 & 12.3\\
        $\mathbb{B}$& 18.3$_{\textcolor{darkblue}{\uparrow\operatorname{11.0}}}$ 
        & 16.6$_{\textcolor{darkblue}{\uparrow\operatorname{11.6}}}$ 
                     & 25.5$_{\textcolor{darkblue}{\uparrow\operatorname{15.7}}}$ 
                     & 31.4$_{\textcolor{darkblue}{\uparrow\operatorname{19.1}}}$\\
        $\mathbb{C}$ & 23.5$_{\textcolor{darkblue}{\uparrow\operatorname{5.2}}}$ 
                     & 9.6$_{\textcolor{darkred}{\downarrow\operatorname{7.0}}}$ 
                     & 12.8$_{\textcolor{darkred}{\downarrow\operatorname{12.7}}}$ 
                     & 43.5$_{\textcolor{darkblue}{\uparrow\operatorname{12.1}}}$\\
        $\mathbb{D}$ & 25.6$_{\textcolor{darkblue}{\uparrow\operatorname{2.1}}}$ 
                     & 17.2$_{\textcolor{darkblue}{\uparrow\operatorname{7.6}}}$ 
                     & 24.4$_{\textcolor{darkblue}{\uparrow\operatorname{11.6}}}$ 
                     & 43.4$_{\textcolor{darkred}{\downarrow\operatorname{0.1}}}$
                     \\
        \bottomrule
    \end{tabular}\hss}}
    \label{tab:lidar_ablation}
\end{minipage}%
\hfill
\begin{minipage}[t]{0.49\textwidth}
    \centering
    
    \includegraphics[width=\linewidth]{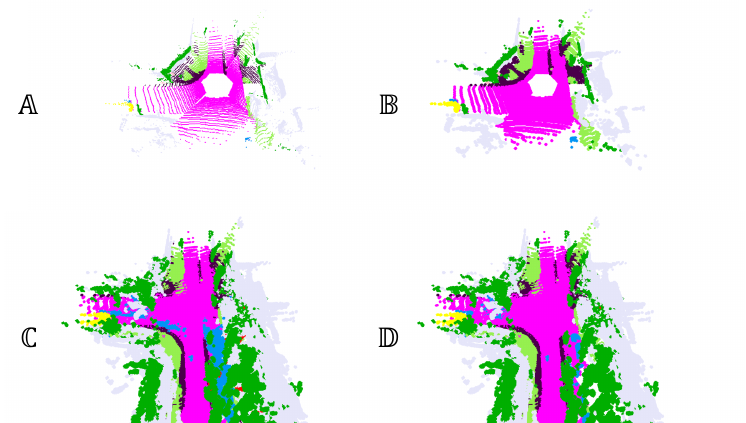}
    \caption*{(b) Ablation Results of $\operatorname{TT\text{-}OccCamera}$}
    % \captionof{table}{Ablation results of $\operatorname{TT\text{-}OccCamera}$.}
    \setlength{\tabcolsep}{9.7pt} 
    \resizebox{0.85\linewidth}{!}{\hbox to 1.0\linewidth{
    \begin{tabular}{c|c c c c}
        \toprule
        No. & mIoU~$\uparrow$
        & \rotatebox{90}{\textcolor{nbus}{$\blacksquare$} $\operatorname{bus}$}
        & \rotatebox{90}{\textcolor{npedestrian}{$\blacksquare$} $\operatorname{ped}$}
        & \rotatebox{90}{\textcolor{nmanmade}{$\blacksquare$} $\operatorname{man}$}\\
        \midrule
        $\mathbb{A}$ & 4.2 & 2.5 & 3.3 & 3.6\\
        $\mathbb{B}$ & 8.5$_{\textcolor{darkblue}{\uparrow\operatorname{4.3}}}$ 
                     & 6.1$_{\textcolor{darkblue}{\uparrow\operatorname{3.6}}}$ 
                     & 5.5$_{\textcolor{darkblue}{\uparrow\operatorname{2.2}}}$ 
                     & 8.3$_{\textcolor{darkblue}{\uparrow\operatorname{4.7}}}$\\
        $\mathbb{C}$ & 14.1$_{\textcolor{darkblue}{\uparrow\operatorname{5.6}}}$ 
                     & 5.6$_{\textcolor{darkred}{\downarrow\operatorname{0.5}}}$ 
                     & 4.7$_{\textcolor{darkred}{\downarrow\operatorname{0.8}}}$ 
                     & 15.3$_{\textcolor{darkblue}{\uparrow\operatorname{7.0}}}$\\
        $\mathbb{D}$ & 14.1$_{\textcolor{darkblue}{\uparrow\operatorname{0.0}}}$ 
                     & 8.0$_{\textcolor{darkblue}{\uparrow\operatorname{2.4}}}$ 
                     & 5.3$_{\textcolor{darkblue}{\uparrow\operatorname{0.6}}}$ 
                     & 15.3$_{\textcolor{darkred}{\downarrow\operatorname{0.0}}}$\\
        \bottomrule
    \end{tabular}\hss}}
    \label{tab:camera_ablation}
\end{minipage}

\caption{\textbf{Qualitative and quantitative comparisons of TT-Occ variants.}
Left: TT-OccLiDAR visualization and ablation results. 
Right: TT-OccCamera visualization and ablation results. 
Each variant incrementally incorporates additional components:
$\mathbb{A}$: Baseline; 
$\mathbb{B}$: Covariance-aware Voxelization; 
$\mathbb{C}$: Inherit Previous Gaussians; 
$\mathbb{D}$: Track Dynamic Gaussians. 
Please zoom in to view details.}
\label{camera_ablation}
% \vspace{-2ex}
\end{figure*}

% Specifically, we achieve this by simply adjusting the resolution during Gaussian voxelization. 
% In contrast, previous methods rely on a fixed resolution predetermined before training and cannot generalize to different resolutions without retraining.
\noindent\textbf{Results on nuCraft} are summarized in Table~\ref{tab:nu}.
As no prior self-supervised methods have been trained or evaluated under this setting, we adapt SelfOcc~\cite{SelfOcc} using its official implementation and checkpoint as a baseline for comparison.
As shown in the table, TT-Occ consistently and significantly outperforms SelfOcc, demonstrating superior adaptability and robustness across \textit{varying resolutions}.

\noindent\textbf{Qualitative comparisons on nuCraft} between both variants of TT-Occ and SelfOcc \cite{SelfOcc} are shown in Fig.~\ref{fig:qualitative}.
Several key observations emerge from these results.
(1) Both our LiDAR and camera variants produce highly accurate occupancy predictions that closely align with the ground truth.
In contrast, SelfOcc generates overly dense predictions, assigning occupancy to nearly all voxels, including empty regions.
This not only incurs significant computational redundancy but also results in severe discrepancies with the ground truth, particularly around dynamic objects (see the radial blue regions).
(2) The LiDAR-based variant produces geometrically accurate reconstructions with broad spatial coverage. However, its fidelity is inherently constrained by the sparsity of LiDAR returns, especially for small or partially scanned objects such as vehicles.
(3) The camera-based variant offers denser reconstructions and better captures small objects within the field of view. Nonetheless, it may struggle with distant regions due to occlusions or limited depth resolution, and the geometry inferred from depth estimation is generally less accurate than that derived from LiDAR.
Despite these challenges, $\operatorname{TT\text{-}OccCamera}$ still remains the state-of-the-art among vision-only occupancy methods.

\subsection{Ablation Studies}
\label{sec:ab}
To evaluate the effectiveness of each component in TT-Occ, we conduct ablation studies on a 10\% held-out validation subset of Occ3D-nuScenes~\cite{Occ3D}.
Since dynamic classes typically occupy only a small portion of the scene but play a critical role in both human perception and downstream tasks, we report not only the overall IoU and mIoU, but also the IoU of representative dynamic classes (bus, pedestrian) and a representative static class (manmade).
We use 3DGS~\cite{3DGS} as the baseline ($\mathbb{A}$), where Gaussians are initialized using the ``lift” strategy introduced in Section~\ref{sec:gaussian} at each time step without temporal information. Gaussians are voxelized by directly scattering their centers. 
As shown in the tables present in Fig. \ref{camera_ablation}, this naïve approach yields poor results due to sparse observations, emphasizing the importance of using anisotropic Gaussian occupancy to better approximate scene geometry.
Next, we introduce covariance-aware voxelization (Eq.~\eqref{eq:voxelization}) and apply sigmoid-based scale regulation ($\mathbb{B}$). 
These lead to consistent improvements across both static and dynamic classes for both LiDAR and camera inputs.

Both Variants $\mathbb{A}$ and $\mathbb{B}$ are single-frame models. 
Allowing Gaussians to directly accumulate across frames without tracking ($\mathbb{C}$) greatly improves the overall and static class performance (\textit{e.g.,} manmade) due to the aggregation of Gaussians for static content, which dominates the scene. However, dynamic class performance drops significantly, as untracked accumulation of moving Gaussians causes temporal inconsistency (see $\mathbb{C}$ in Fig.~\ref{fig:trail} for trailing artifacts).
To address this, we incorporate tracking dynamic Gaussians as described in Section~\ref{sec:flow}, which significantly improves the accuracy of dynamic classes while maintaining performance on static content. 
As shown in $\mathbb{D}$, this yields cleaner occupancy with trailing and ghosting artifacts largely eliminated.
The ablation study on the optional TRBF fusion module is presented in supplementary materials.

\begin{table}[t]
\centering
\caption{Impact of semantic and geometric VFMs of $\mathrm{TT\text{-}Occ}$.}
\vspace{-2ex}
\resizebox{\linewidth}{!}{
\begin{tabular}{lccc}
\toprule
\textbf{Method} & \textbf{Semantic} & \textbf{Geometry} & \textbf{mIoU}~$\uparrow$ \\
\midrule
$\mathrm{SelfOcc}$ & $\mathrm{OpenSeeD}$~\cite{OpenSeeD} & --  & 11.6 \\
\midrule
\multirow{3}{*}{$\mathrm{TT\text{-}OccLiDAR}$}  & $\mathrm{OpenSeeD}$~\cite{OpenSeeD} & -- & 23.6 \\
&  $\mathrm{GroundedSAM2}$~\cite{groundedsam} & -- & 21.3 \\
 & $\mathrm{REX\text{-}Omni}$~\cite{rexomni} & -- & \textbf{27.4} \\
\midrule
% $\mathrm{TT\text{-}Occ}$ & Camera & $\mathrm{OpenSeeD}$~\cite{OpenSeeD} & $\mathrm{R3D3}$~\cite{r3d3} & 14.0 \\
\multirow{5}{*}{$\mathrm{TT\text{-}OccCamera}$}  & $\mathrm{OpenSeeD}$~\cite{OpenSeeD} & $\mathrm{VGGT}$~\cite{vggt} & 13.4 \\
  & $\mathrm{OpenSeeD}$~\cite{OpenSeeD} & $\mathrm{MapAnything}$~\cite{keetha2025mapanything} & 14.3 \\
& $\mathrm{GroundedSAM2}$~\cite{groundedsam} & $\mathrm{VGGT}$~\cite{vggt} & 12.2 \\
 & $\mathrm{REX\text{-}Omni}$~\cite{rexomni} & $\mathrm{VGGT}$~\cite{vggt} & 15.9 \\
 & $\mathrm{REX\text{-}Omni}$~\cite{rexomni} & $\mathrm{MapAnything}$~\cite{keetha2025mapanything} & \textbf{16.7} \\
\bottomrule
\end{tabular}
}
\label{tab:semantic_geometry}
% \vspace{-2ex}
\end{table}

\noindent \textbf{Ablation on Loosely Coupled VFM Components.}
To demonstrate that our framework is loosely coupled and can support arbitrary VFMs, and to further investigate how different VFMs affect overall performance, we integrate three semantic models, namely OpenSeeD \cite{OpenSeeD}, GroundingSAM2 \cite{groundedsam}, and REX-Omni \cite{rexomni}, as well as two depth-estimation backbones, VGGT \cite{vggt} and MapAnything \cite{keetha2025mapanything}. 
Results of all combinations on Occ3D-nuScenes \cite{Occ3D} are summarized in Table~\ref{tab:semantic_geometry}. 
GroundingSAM2 yields the weakest performance, while REX-Omni provides the strongest semantic cues. 
Among depth models, MapAnything consistently outperforms VGGT, primarily because it provides metric-scale depth directly, whereas VGGT requires additional triangulation to recover metric scale, which inevitably introduces errors. 
Overall, these experiments confirm that our loosely coupled, modular design allows seamless integration of advanced VFMs without modification, directly benefiting from their rapid evolution and scaling trend.

\begin{figure}
    \centering
    \includegraphics[width=0.9\linewidth]{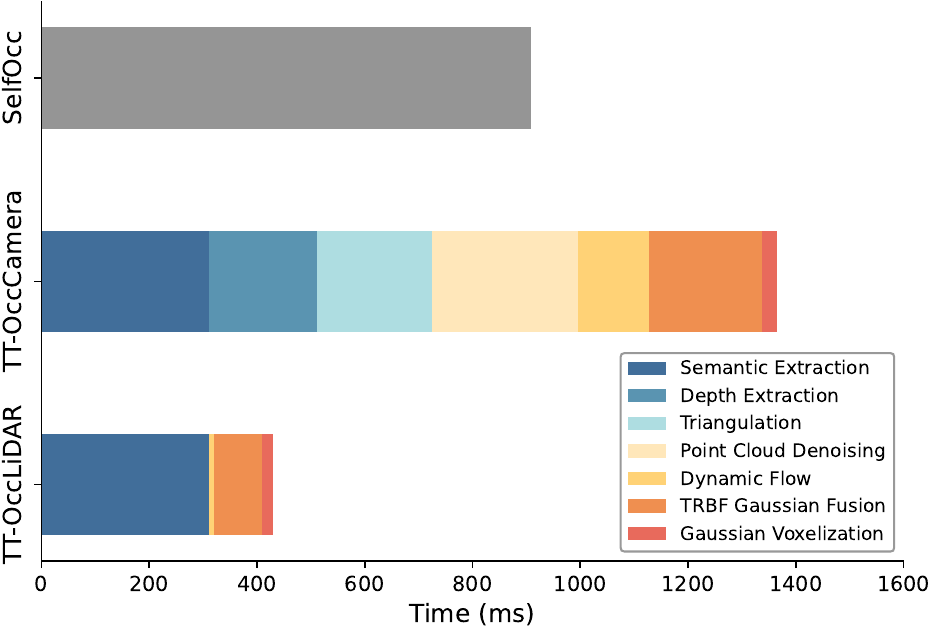}
    % \vspace{-20pt}
    \caption{Inference time comparison among TT-OccCamera, TT-OccLiDAR, and SelfOcc \cite{SelfOcc}. The horizontal stacked bars illustrate the per-module runtime composition of each module.}
    % \vspace{-2ex}
    \label{fig:infertime}
\end{figure}

\noindent \textbf{Efficiency Analysis.} We provide a detailed runtime breakdown of our pipeline for the vision-centric and LiDAR-based variants in Fig. \ref{fig:infertime} (using OpenSeeD \cite{OpenSeeD}). The reported values represent the average processing time per timestep across six input images.
Semantic segmentation using OpenSeeD~\cite{OpenSeeD} constitutes the most computationally intensive step in both pipelines, accounting for 28.5\% of total runtime in the camera variant and 77.9\% in the LiDAR variant. 
In the vision-centric scenario, the absence of LiDAR data requires additional processes such as depth estimation, triangulation-based calibration, and point cloud denoising, collectively contributing 46.5\% of the overall runtime.
Gaussian voxelization and the optional TRBF fusion module are relatively efficient; however, their runtime is proportional to the number of Gaussians involved. Therefore, the camera-based pipeline which has denser Gaussians experiences slightly increased computational overhead compared to its LiDAR-based counterpart.
Finally, tracking dynamic Gaussians in $\operatorname{TT\text{-}OccCamera}$ incurs much higher computational costs compared to $\operatorname{TT\text{-}OccLiDAR}$ due to its reliance on dense optical flow estimation across six images using RAFT~\cite{raft}, whereas the LiDAR variant only applies lightweight ICP alignment for sparse foreground points.

We report the average peak GPU memory usage on high-resolution nuCraft \cite{nuCraft} in Table~\ref{tab:peak_memory}.
Because our system processes modules sequentially, the peak memory footprint is determined by the most demanding component at runtime (\textit{e.g.}, OpenSeeD or the accumulated Gaussians). In the LiDAR variant, memory usage is primarily dominated by the semantic module, whereas in the camera variant, the growing set of Gaussians becomes the major contributor.
Overall, TT-Occ maintains a peak memory consumption below 10GB for both LiDAR and camera settings.

% Other components shared by both variants, such as TRBF Gaussian fusion and voxelization, are highly efficient and together account for less than 15\% of the total runtime in both cases, demonstrating the lightweight nature of our core pipeline.

\begin{table}
\centering
\setlength{\tabcolsep}{3.5pt}
\renewcommand{\arraystretch}{1.05}
\caption{Memory statistics of TT-Occ.}\vspace{-2ex}
\resizebox{0.82\columnwidth}{!}{
\begin{tabular}{lcc}
\toprule
\textbf{Variant} & \textbf{Gaussians} & \textbf{Peak Mem (GB)} \\
\midrule
$\mathrm{TT\text{-}OccLiDAR}$& 786{,}874 & 5.6 \textit{(dom. by OpenSeeD)} \\
$\mathrm{TT\text{-}OccCamera}$ & 2{,}968{,}647 & 9.9 \textit{(dom. by Gaussians)} \\
\bottomrule
\end{tabular}}
\label{tab:peak_memory}
% \vspace{-2ex}
\end{table}

\section{Conclusion}
% In this work, we introduce TT-Occ, an efficient and reliable method for occupancy prediction that removes the requirement for expensive pre-training and supports dynamic, real-time adaptation to varying object classes and spatial resolutions. Extensive experiments conducted on Occ3D-nuScenes and nuCraft demonstrate both quantitatively and qualitatively that TT-Occ effectively scales to higher resolutions, achieving up to a 45\% improvement in mIoU. Nevertheless, sparse initialization of Gaussians can cause inaccuracies in Gaussian flows and lifted semantics, which are difficult to correct without explicit 3D supervision. Although trilateral smoothing partially mitigates this issue, accumulating point clouds across multiple sweeps could further densify Gaussians and improve reconstruction quality at the expense of higher inference-time costs. Future work will explore methods to balance reconstruction accuracy and computational efficiency.

In this paper, we presented TT-Occ, a practical and flexible test-time 3D occupancy framework that turns generic vision foundation models into an effective occupancy predictor through time-aware 3D Gaussians. By bypassing conventional dense occupancy decoders, TT-Occ enables arbitrary voxel resolutions, open-vocabulary object recognition, and strong temporal coherence without any additional pretraining. Extensive evaluations on Occ3D and nuCraft demonstrate that both our LiDAR-based and vision-centric variants deliver consistent accuracy gains and competitive efficiency, underscoring the promise of TT-Occ as a deployable, VFM-native solution for real-world driving environments.

\section{Acknowledgment}
This work was partially supported by ARC DE240100105, DP240101814, DP230101196, BA24006, and ARC Industrial Transformation Research Hubs IH230100013.

{
    \small
    \bibliographystyle{ieeenat_fullname}
    \bibliography{main}
}

\clearpage
\setcounter{page}{1}
\maketitlesupplementary

\begin{figure*}[t]
    \centering
    \begin{subfigure}{0.95\linewidth}
        \centering
        \includegraphics[width=1\linewidth]{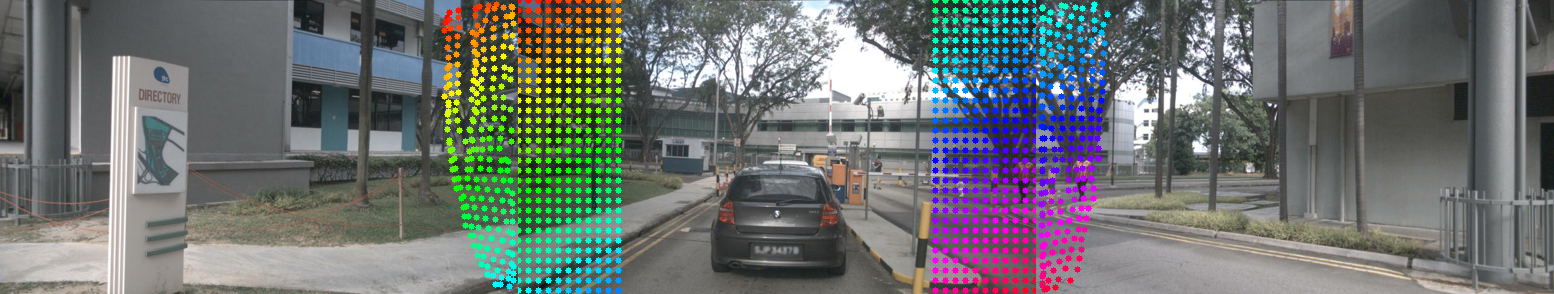}
    \end{subfigure}
    
    % \vspace{-1mm} % 调整上下间距（可微调为 -1mm ~ -4mm）

    % \begin{subfigure}{1.0\linewidth}
    %     \centering
    %     \includegraphics[width=1.0\linewidth]{Figs/2Dtrack2.png}
    % \end{subfigure}

    \begin{subfigure}{0.95\linewidth}
        \centering
        \includegraphics[width=1\linewidth]{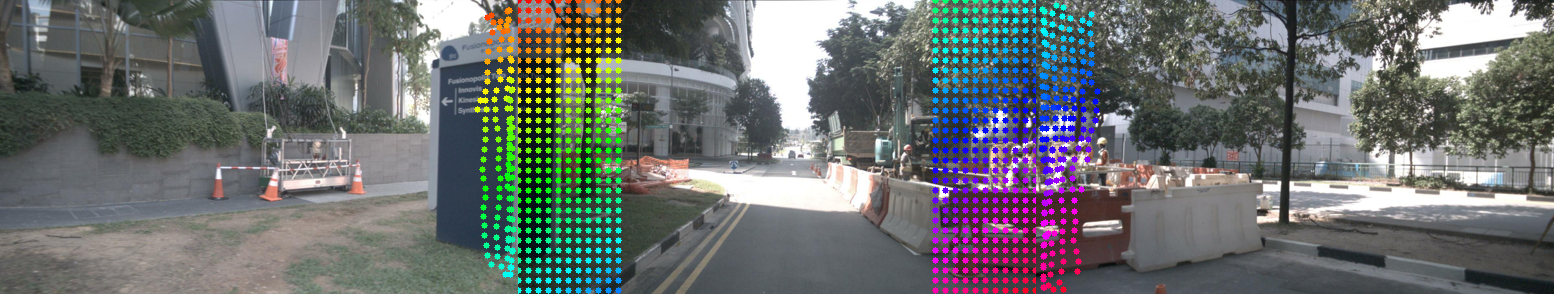}
    \end{subfigure}
    
    \caption{Visualization of VGGT-predicted 2D tracking across front-left, front, and front-right cameras. Sparse query points are tracked and subsequently triangulated to obtain a metric 3D point cloud, which is used to align the predicted depth maps to 
    real-world scale.
    }
    \label{fig:2d_track}
\end{figure*}

\appendix

\section{Implementation Details}
% In this Appendix, we provide implementation details and additional visualizations. 

%%%%%%%%%%%%%%%%%%%%%%%%%%%%%%%%%%%%%%%%%%%%%%%%%%%%%%%%%%%%
% \subsection{Implementation Details}\label{sec:impl-Camera}
% In this section, we provide a detailed description of the technical components of our system.

\subsection{Depth Estimation and Triangulation-Based Calibration for TT-OccCamera}
\label{sup:vggt}
% \begin{figure*}[!htb]
%     \includegraphics[width=1\linewidth]{Figs/03_tracks_grid.png}
%     \caption{Visualization of VGGT-predicted 2D tracking across front-left, front, and front-right camera views. Sparse query points are tracked across multiple views and subsequently triangulated to obtain a metric 3D point cloud, which is used to align the predicted depth maps to real-world scale.}
%     \label{fig:2d_track}
% \end{figure*}

\begin{figure*}[ht!b]
\centering
    \includegraphics[width=0.95\linewidth]{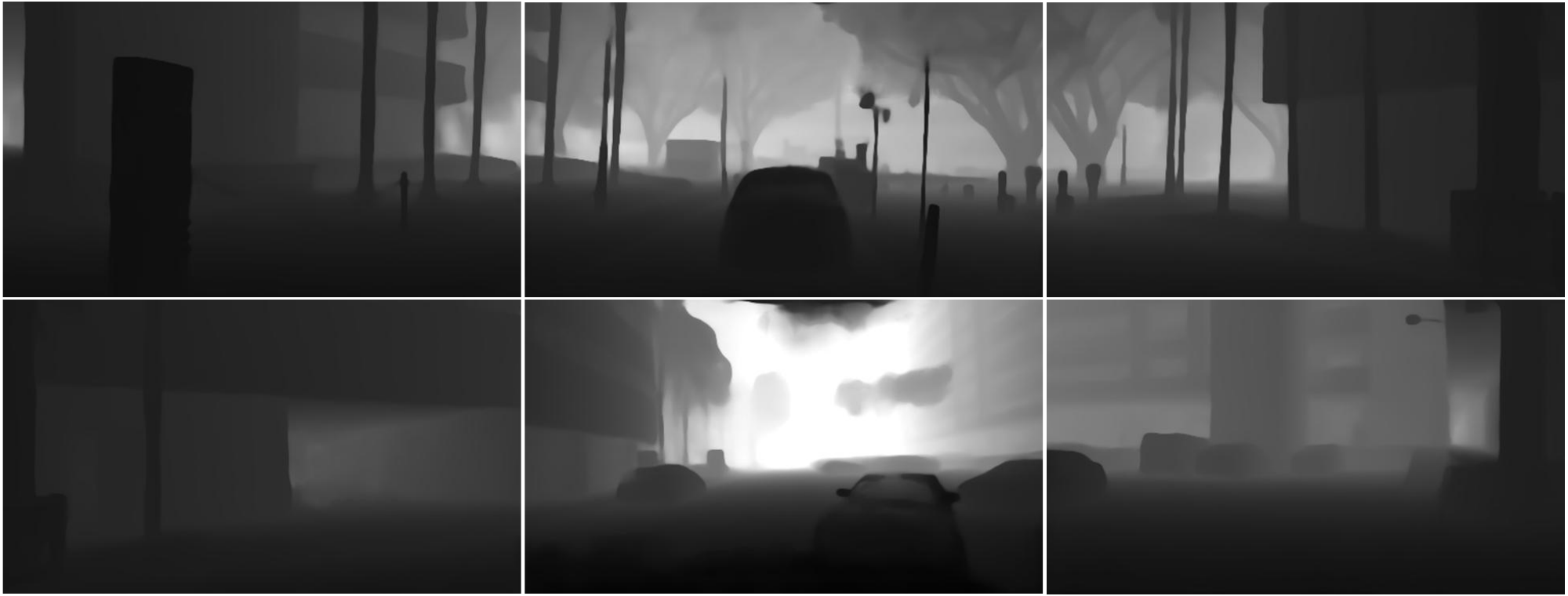}
    \caption{Visualization of scaled VGGT depth prediction on example frames.}
    \label{fig:depth}
\end{figure*}

\begin{figure*}[ht!b]
\centering
    \includegraphics[width=0.95\linewidth]{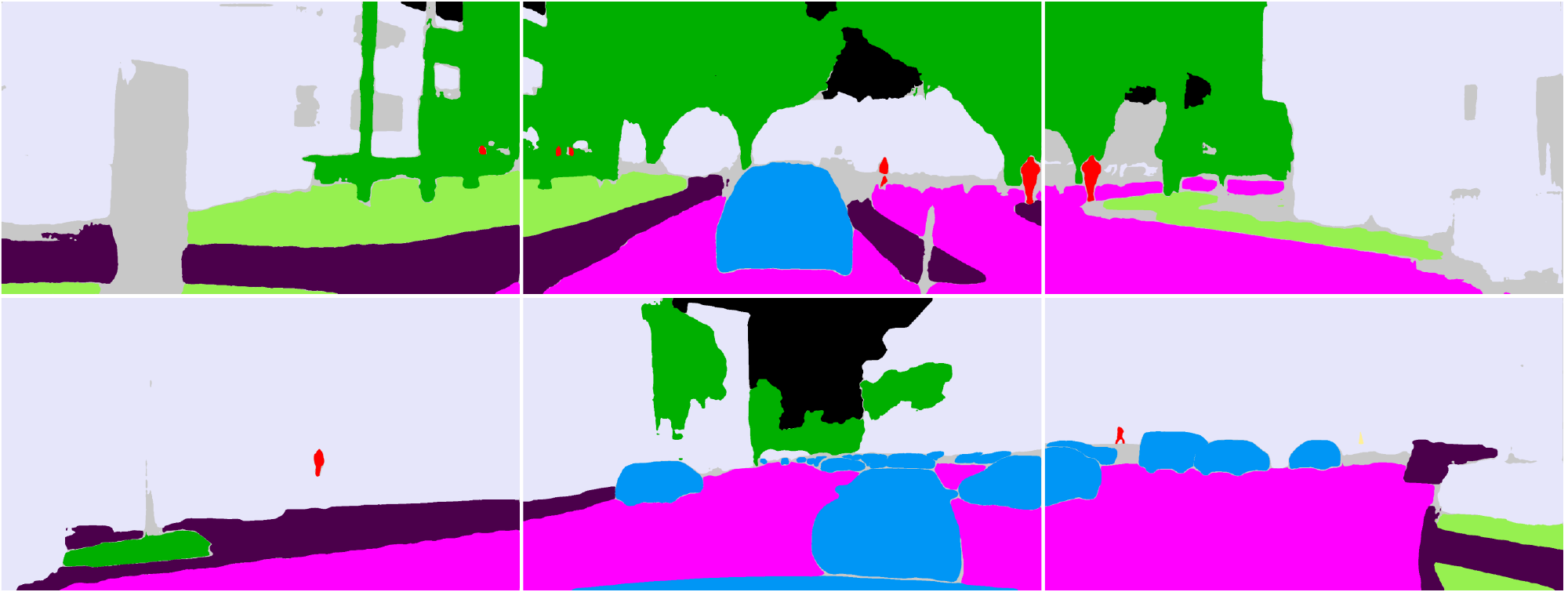}
    \caption{Visualization of OpenSeeD segmentation results on example frames.\vspace{-1ex}}
    \label{fig:openseed}
\end{figure*}

When the outputs of 3DVFMs are already in a metric scale (such as MapAnything \cite{keetha2025mapanything}, they can be directly applied. 
However, for models that do not provide metric outputs (such as VGGT \cite{vggt}), additional calibration is required to align them with the ground-truth occupancy. 
Using VGGT as an example, we describe below the depth estimation process and a triangulation-based calibration method.
VGGT is a feed-forward neural network capable of predicting depth maps and tracking 2D keypoints across frames.
We input six surrounding camera views into VGGT to generate per-view depth predictions. 
Following the original VGGT setup, the input images are resized to a resolution of 294~$\times$~518.
Although VGGT produces consistent and high-quality depth estimates across views, the predictions are in an unscaled unit space and do not correspond directly to real-world metric distances.
To address this limitation, we leverage VGGT’s built-in 2D point tracking functionality across multiple views at the same time step. 
Specifically, we select three adjacent cameras including front, front-left, and front-right, and use VGGT to track sparse 2D keypoints across them.
By filtering out low-quality matches using the predicted visibility and confidence scores, we obtain reliable point correspondences between camera pairs, as illustrated in Fig.~\ref{fig:2d_track}.
We then triangulate these matched 2D points using the ground-truth camera intrinsics and extrinsics provided by the dataset, resulting in a sparse but metrically accurate 3D point cloud.
Finally, we compare the magnitudes of the triangulated 3D points with those reconstructed from the predicted depth maps at the corresponding image locations, and compute a global scaling factor to align the depth predictions with real-world scale.
An example of the final scaled depth prediction is shown in Fig.~\ref{fig:depth}.

\subsection{Open-Vocabulary Semantic Segmentation}
\label{sup:openseed}
We now describe the process of open-vocabulary semantic segmentation using OpenSeeD \cite{OpenSeeD} as an example. 
We adopt OpenSeeD primarily to ensure a fair comparison with SelfOcc~\cite{SelfOcc}. 
Nevertheless, our system is loosely coupled with VFMs and fully compatible with more advanced ones, such as REX-Omni \cite{rexomni}.
As shown in Fig. \ref{fig:openseed}, OpenSeeD's predictions often exhibit noisy and unclear boundaries. 
% This issue becomes even more evident when projecting the results into 3D space. 
% We adopt OpenSeeD \cite{OpenSeeD}, a simple and early framework for open-vocabulary segmentation, to extract semantic information from six surrounding images at each timestamp. 
% We choose to use OpenSeeD primarily to ensure a fair comparison with SelfOcc \cite{SelfOcc}, but it is important to note that our pipeline is loosely coupled with VFMs and any model capable of open-vocabulary segmentation can be seamlessly integrated into our system. 
% We plan to support more advanced segmentation models in our future open-source release.
% Although OpenSeeD \cite{OpenSeeD} accounts for a significant portion of the total runtime, we choose to feed it with full-resolution images since we observed that OpenSeeD is sensitive to image resolution: downsampling leads to noticeable degradation in segmentation accuracy, especially for small objects.
% The prompt words we use include: ``bicycle'', ``bus'', ``car'', ``sedan'', ``van'', ``construction vehicle'', ``crane'', ``excavator'', ``motorcycle'', ``person'', ``pedestrian'', ``truck'', ``traffic cone'', ``cone'', ``road'', ``highway'', ``street'', ``sidewalk'', ``terrain'', ``grass'', ``building'', ``wall'', ``fence'', ``bridge'', ``pole'', ``traffic pole'', ``traffic light'', ``traffic sign'', ``street sign'', ``street pole'', ``streetlight'', ``hydrant'', ``meter box'', ``display window'', ``skyscraper'', ``parking meter'', ``tower'', ``house'', ``structure'', ``banner'', ``board'', ``billboard'', ``stairs'', ``pillar'', ``tree'', and ``sky''.
Since our focus is not prompt engineering and to ensure a fair comparison with SelfOcc \cite{SelfOcc},
we adopt the same query set including: \textit{"barrier"}, \textit{"bicycle"}, \textit{"bus"}, \textit{"car"}, \textit{"construction\_vehicle"}, \textit{"crane"}, \textit{"motorcycle"}, \textit{"person"}, \textit{"traffic\_cone"}, \textit{"trailer"}, \textit{"trailer\_truck"}, \textit{"truck"}, \textit{"road"}, \textit{"sidewalk"}, \textit{"terrain"}, \textit{"grass"}, \textit{"building"}, \textit{"wall"}, \textit{"tree"}, \textit{"sky"}.

\begin{figure*}[ht!]
    \includegraphics[width=0.87\linewidth]{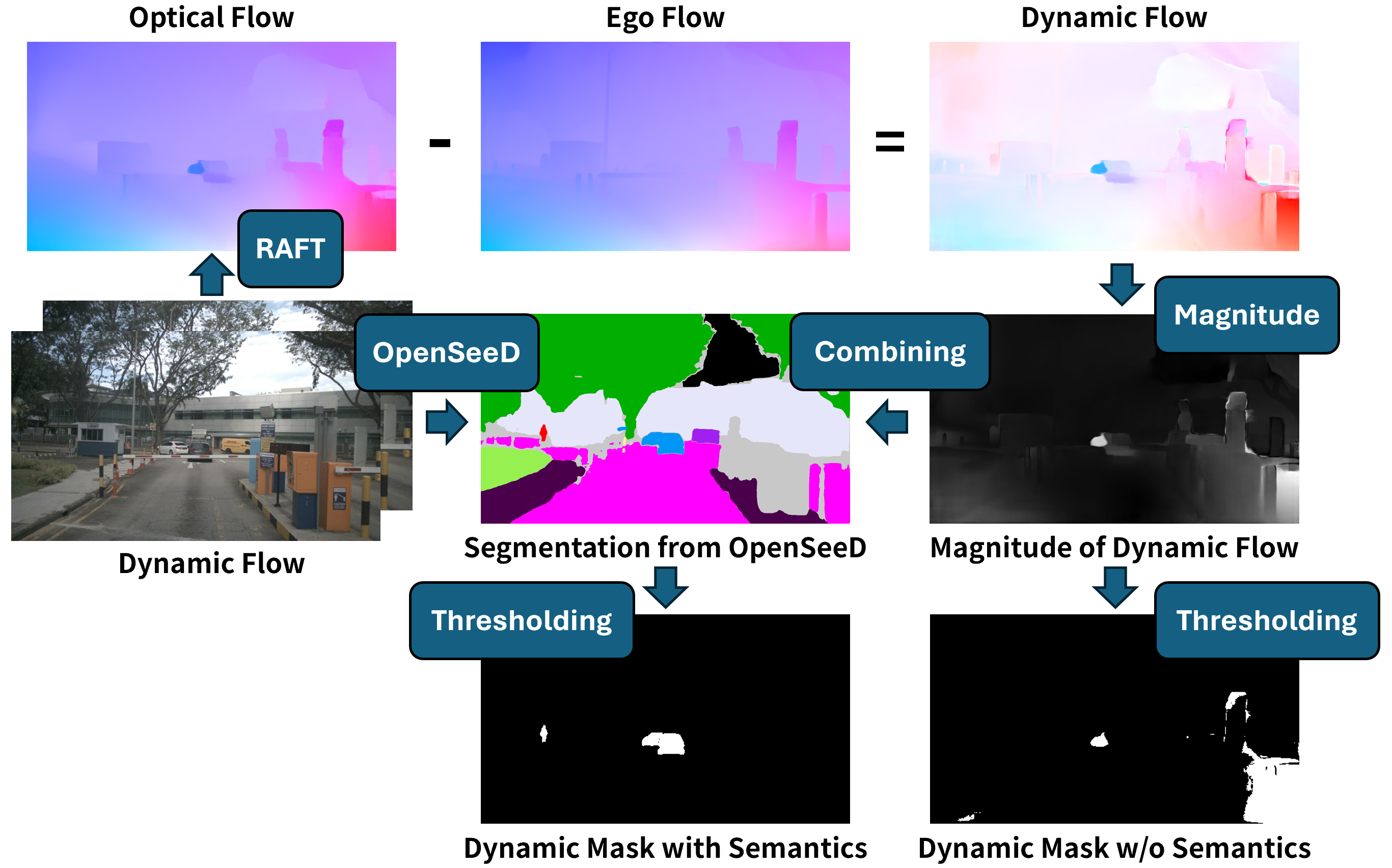}
    \centering
    \captionof{figure}{Illustration of the tracking process in $\operatorname{TT\text{-}OccCamera}$. \vspace{-1ex}}
    \label{fig:raft}
\end{figure*}

\subsection{Tracking with RAFT for TT-OccCamera}
\label{sup:depth}
For $\operatorname{TT\text{-}OccCamera}$, we estimate the optical flow $F_{\text{opt}}$ between two consecutive frames from the same camera using RAFT \cite{raft}.
We then compute the ego-motion-induced flow $F_{\text{ego}}$ based on the ground-truth camera intrinsics and extrinsics of the adjacent frames, along with the predicted depth from 3DVFMs.
By subtracting the ego flow from the observed optical flow, we obtain the dynamic flow $F_{\text{dyn}} = F_{\text{opt}} - F_{\text{ego}}$, which theoretically captures the motion of dynamic objects in the environment.
Although this 2D dynamic flow could, in principle, guide the 3D motion of dynamic Gaussians, back-projecting it into 3D space tends to amplify errors from RAFT and 3DVFMs, resulting in unstable Gaussian motion.
To mitigate this, we adopt a compromise strategy by thresholding the dynamic flow magnitude to obtain a dynamic mask that identifies likely moving regions. 
In the ideal case, a simple thresholding on the magnitude of $F_{\text{dyn}}$ would yield a reliable binary mask for dynamic regions.
However, since both $F_{\text{opt}}$ and $F_{\text{ego}}$ are derived from 2D estimations and are subject to noise and inaccuracies, the resulting $F_{\text{dyn}}$ is often highly unreliable and noisy.
To further refine the dynamic flow, we leverage the the cues from semantic segmentation models, which provides relatively cleaner object boundaries, to refine the dynamic flow magnitude map.
As illustrated in Fig. \ref{fig:raft}, the raw dynamic flow is noisy, and thresholding it directly often produces fragmented masks that do not correspond to coherent objects.
After incorporating instance masks from segmentation models, high-magnitude errors on the background are suppressed, and the resulting dynamic masks become more object-aligned, either an entire object is identified as dynamic or it is not, effectively eliminating partial or spurious activations.
The corresponding 3D Gaussians projected onto these regions are treated as dynamic and excluded from static accumulation in the next frame. While this approach does not allow accumulation of dynamic objects as in the LiDAR-based variant, it effectively reduces artifacts caused by noisy motion cues and temporal inconsistencies.

\subsection{Tracking with LiDAR for TT-OccLiDAR}
\label{sup:lidar}
\begin{figure}
    \includegraphics[width=1\linewidth]{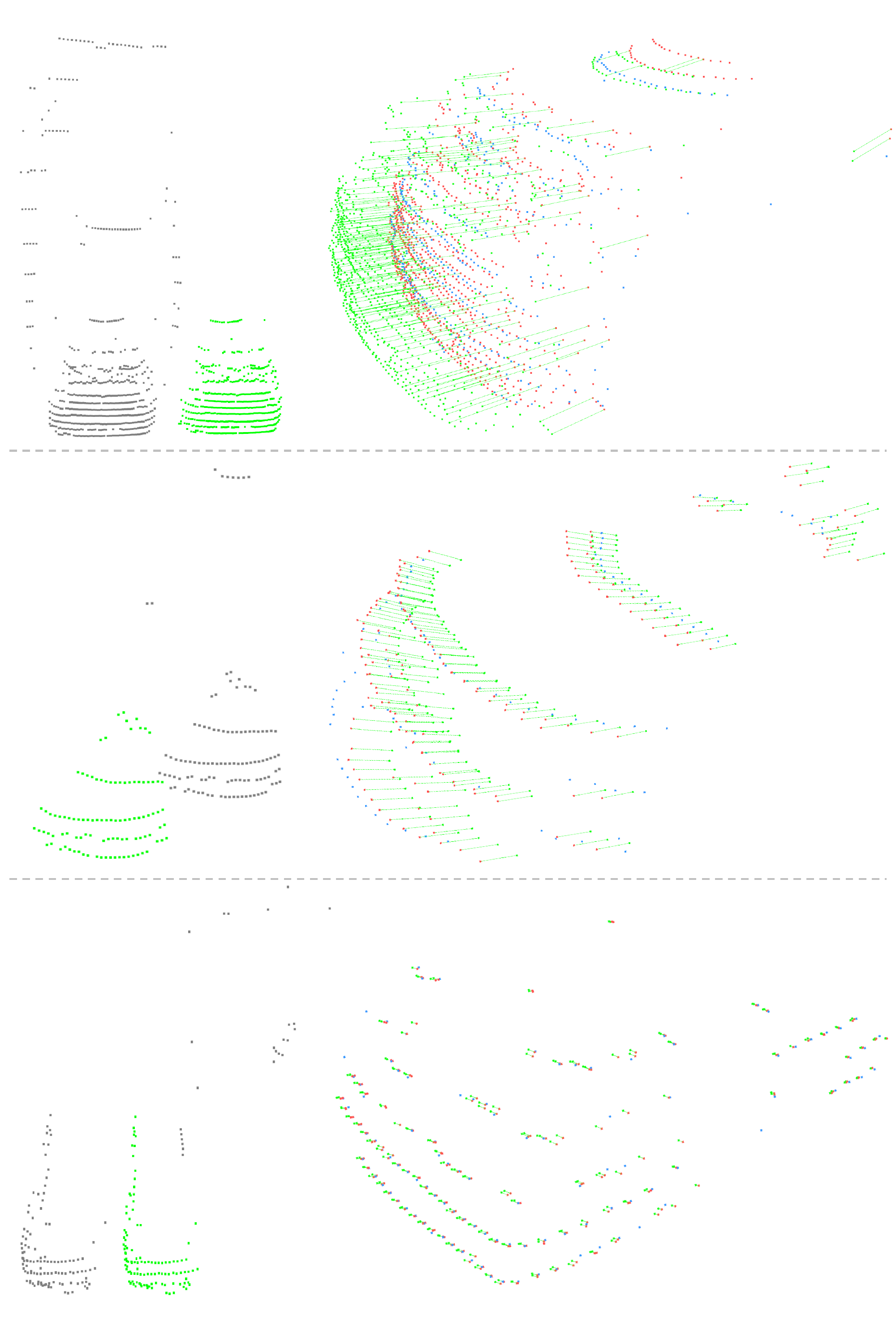}
    \centering
    \captionof{figure}{\textbf{Visualization of instance-level point cloud denoising and 3D flow estimation.}
From the first to the third row, we show three example \textbf{car} instances from LiDAR data. 
Left: gray points represent the raw instance points obtained from segmentation masks, while green points are the core structures extracted via DBSCAN (visualized with a slight offset for clarity). 
Right: ICP-estimated 3D flow between adjacent frames, where green, blue, and red denote the source, target, and aligned source points, respectively. Green lines indicate the estimated flow vectors. 
It can be seen that DBSCAN effectively removes noisy outliers, and ICP produces accurate frame-to-frame alignment.
}
    \label{fig:lidar}
\end{figure}

Tracking in $\operatorname{TT\text{-}OccLiDAR}$ is generally more reliable than in $\operatorname{TT\text{-}OccCamera}$, as LiDAR point clouds provide more accurate and consistent geometric information.
We follow a straightforward strategy: cluster first, then align via ICP.
First, we project LiDAR points onto the instance masks predicted by the segmentation model, thereby associating each point with a specific foreground object.
Due to the often imprecise boundaries of predicted masks, the resulting instance-level point sets can contain substantial noise. To address this, we apply DBSCAN clustering \cite{dbscan} to each instance’s point cloud to extract its core structure and eliminate outliers. This approach proves effective in significantly removing noise, as illustrated in the left column of Fig.~\ref{fig:lidar}, where gray points are obtained by directing projecting onto segmentation masks and green points represent the denoised output after DBSCAN clustering (slightly translated for observation).
We then perform object-level matching across adjacent frames based on the spatial proximity and shape similarity of the filtered point clusters. For each matched pair, the 3D flow is estimated using the Iterative Closest Point (ICP) algorithm \cite{ICP}. Qualitative results are presented in the right column of Fig.~\ref{fig:lidar}, where green, blue, and red points represent the source points, destination points, and the ICP-transformed source points, respectively. Green arrows indicate the estimated 3D flow vectors. The effectiveness of the ICP-based alignment can be clearly observed.
Finally, matched points are propagated to the next frame, while unmatched instances from the previous frame are discarded to avoid the accumulation of errors caused by moving or disappearing objects.

\section{Additional Results}

\subsection{Performance under Challenging Conditions}
A key strength of our system is that it is built upon large-scale foundation models, which are trained on diverse datasets and therefore exhibit strong generalization beyond specific domains. 
Leveraging these models gives TT-Occ a natural advantage over task-specific approaches such as SelfOcc~\cite{SelfOcc}, particularly under challenging conditions.
To assess robustness, we evaluated both the camera- and LiDAR-based variants on nighttime and rainy scenes from the nuScenes~\cite{nuScenes} dataset. 
As shown in Table~\ref{tab:cha}, TT-Occ consistently surpasses SelfOcc across all challenging scenarios, demonstrating improved accuracy and robustness.

\begin{table}
\centering
\caption{\textbf{Robustness evaluation on rainy and nighttime nuScenes \cite{nuScenes} scenes (mIoU).} R: rainy. N: nighttime. 
TT-Occ variants consistently outperform the task-specific SelfOcc \cite{SelfOcc} model across all challenging conditions.}
\setlength{\tabcolsep}{6pt}
\renewcommand{\arraystretch}{1.15}
\resizebox{1\linewidth}{!}{
\begin{tabular}{l|ccccc|c}
\toprule
 & \multicolumn{5}{c|}{\textbf{Scene ID}} & \\
\textbf{Method} & \textbf{0911} & \textbf{0915} & \textbf{1065} & \textbf{1067} & \textbf{1073} & \textbf{Avg} \\
 & (R) & (R) & (R+N) & (R+N) & (N) & \\
\midrule
SelfOcc & 18.3 & 11.0 & 7.9 & 7.3 & 6.2 & 10.1 \\
$\mathrm{TT\text{-}OccCamera}$ & 21.2 & 13.6 & 11.2 & 9.4 & 10.3 & 13.1 \\
$\mathrm{TT\text{-}OccLiDAR}$ & \textbf{27.0} & \textbf{18.9} & \textbf{12.0} & \textbf{13.9} & \textbf{16.2} & \textbf{17.6} \\
\bottomrule
\end{tabular}
}
\label{tab:cha}
\end{table}


\subsection{Expanded Ablation Comparisons}

\begin{figure*}[ht!]
    \centering
    \includegraphics[width=0.82\linewidth]{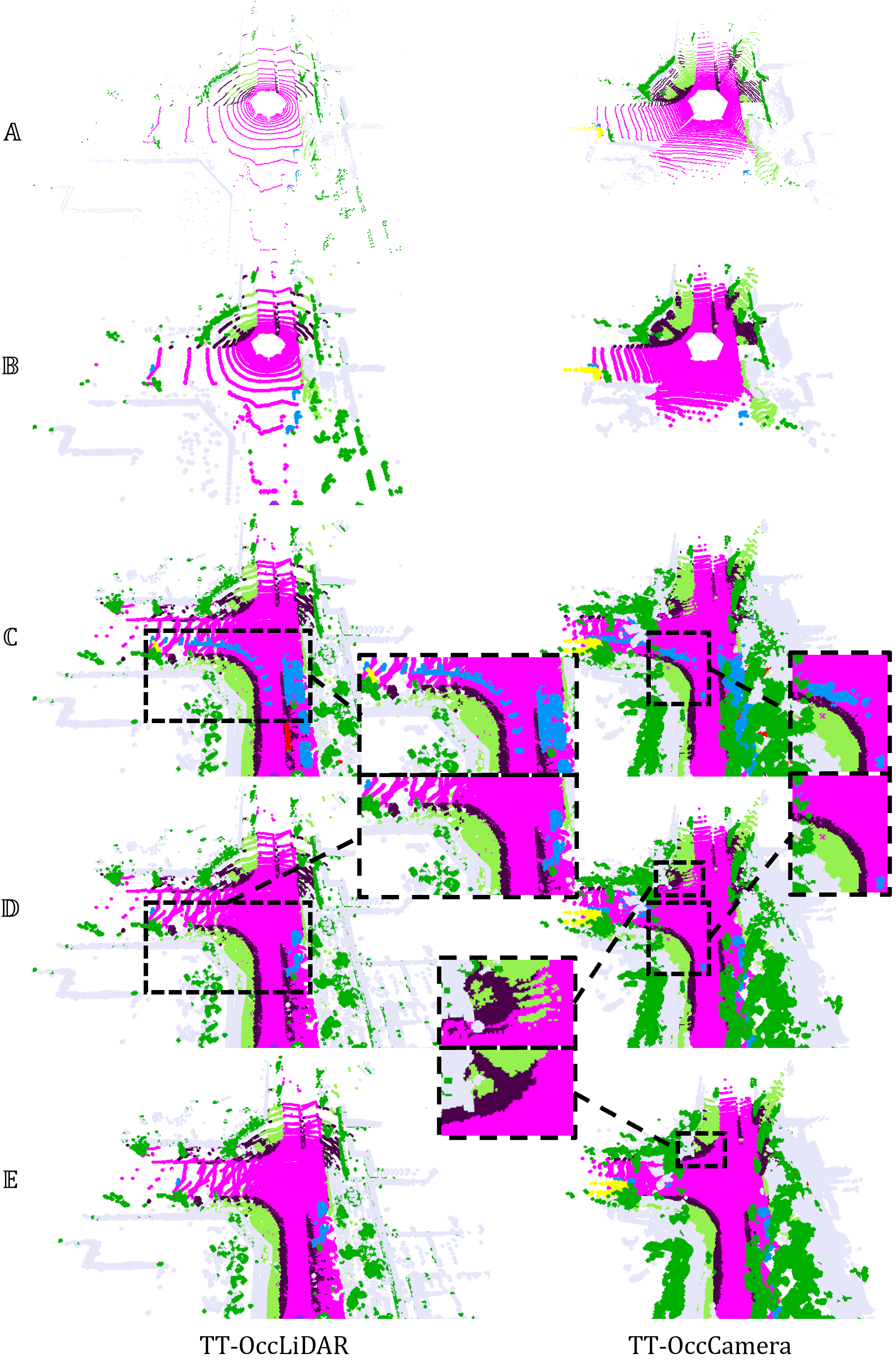}
    \captionof{figure}{Zoomed-in visualization of different baselines of both variants of TT-Occ. $\mathbb{A}$: Baseline. $\mathbb{B}$: Covariance-aware Voxelization. $\mathbb{C}$: Inherit Previous Gaussians. $\mathbb{D}$: Track Dynamic Gaussians. $\mathbb{E}$: TRBF Fusion. 
    }
    \label{supfig:abcde}
\end{figure*}

We further present enlarged visual comparisons of the variants evaluated in our ablation study. 
These visualizations reinforce the conclusions drawn in the main paper. 
Variant $\mathbb{A}$, where Gaussians are initialized using the ``lift” strategy at each time step without temporal information, performs poorly due to sparse observations and the lack of anisotropic occupancy modeling needed to approximate local geometry. 
Introducing covariance-aware voxelization and scale regularization in $\mathbb{B}$ leads to consistent improvements across both static and dynamic classes for both LiDAR and camera settings. 
Allowing Gaussians to accumulate over time in $\mathbb{C}$ further boosts performance on static classes by aggregating evidence across frames, but severely degrades dynamic class accuracy due to untracked motion, resulting in trailing artifacts. 
Incorporating dynamic Gaussian tracking in $\mathbb{D}$ restores temporal consistency and substantially improves dynamic class performance while preserving strong performance on static content, producing clean and artifact-free occupancy.

In addition to these variants, we include baseline $\mathbb{E}$, which integrates the optional TRBF fusion module. 
Although $\mathbb{D}$ already handles dynamic objects effectively, we still observe scattered high-frequency noise, particularly in the camera variant, primarily due to segmentation boundary inaccuracies and imperfect dynamic region estimation. 
While this noise is extremely sparse and has negligible impact on overall accuracy, it slightly reduces visual quality. 
To mitigate this, TRBF fusion is applied as an optional spatio-temporal smoothing module. 
As shown in Fig.~\ref{supfig:abcde}, TRBF in $\operatorname{TT\text{-}OccCamera}\mathbb{E}$ effectively suppresses residual noise and produces smoother and more visually coherent reconstruction results.

\end{document}